\documentclass[11pt]{article}

\usepackage[margin=1in]{geometry}
\usepackage[T1]{fontenc}
\usepackage{lmodern}
\usepackage{amsmath,amssymb,amsthm}

\allowdisplaybreaks[2]

\usepackage{bm}
\usepackage{graphicx}
\usepackage{booktabs}
\usepackage{xcolor}
\usepackage{comment}
\usepackage{algorithm}
\usepackage{algpseudocode}
\usepackage{tikz}
\usepackage[authoryear,round]{natbib}
\usepackage[hidelinks]{hyperref}
\usepackage{setspace}
\newtheorem{prop}{Proposition}
\theoremstyle{remark}
\newtheorem{remark}{Remark}

\title{Prescriptive Learning of Linear Policies with Feasibility Guarantees for Binary Decision-Making}

\author{%
Farzaneh Pourahmadi\thanks{Department of Wind and Energy Systems, Technical University of Denmark. Email: \texttt{farpour@dtu.dk}}
\and
Trine Krogh Boomsma\thanks{Department of Mathematical Sciences, University of Copenhagen. Email: \texttt{trine@math.ku.dk}}
\and
Pierre Pinson\thanks{Dyson School of Design Engineering, Imperial College London. Email: \texttt{p.pinson@imperial.ac.uk}}
}

\date{}

\begin{document}

\maketitle

\begin{abstract}
Binary decision-making under uncertainty arises across a wide range of applications, including energy systems scheduling, transportation, and workforce planning. While machine learning has been increasingly applied to accelerate or replace mixed-integer solvers in such settings, existing approaches rely on predictive learning, training models to reproduce optimal decisions from labeled data, which lacks knowledge of counterfactual outcomes and provides no guarantees that the resulting decisions will be feasible. This limitation is further compounded under distributional shift, where policies trained on historical data may produce systematically infeasible decisions. We propose a prescriptive learning framework for binary decision-making that addresses both limitations. Rather than predicting optimal labels, the framework jointly optimizes over linear decision policies and feasible assignments across all training scenarios, allowing feasible alternatives and their operational performance to be considered directly during training. We further extend the framework to a distributionally robust chance-constrained formulation using a Wasserstein ambiguity set, which enforces feasibility with a prescribed probability under the worst-case distribution in the ambiguity set and enables a tunable trade-off between robustness and solution quality. We demonstrate the framework on two case studies, namely a unit commitment problem and a staff allocation problem. Out-of-sample experiments under test distributions that either match or differ from the training distribution show that the robust model is more conservative when the two distributions coincide, while under distributional shift it substantially reduces unmet demand and lowers total cost by up to 76.5$\%$ relative to classical predictive learning.

\end{abstract}



\section{Introduction}\label{sec:Intro}

Mixed-integer optimization plays a critical role across a wide range of domains, including energy systems, transportation, scheduling, supply chains, and finance. However, solving mixed-integer optimization problems can be computationally challenging, particularly when decisions must be obtained frequently or in real time. Since binary decisions are often a primary source of computational complexity and can be particularly difficult to learn accurately while maintaining feasibility, in this paper, we focus specifically on the binary component of mixed-integer optimization problems.

Machine learning has increasingly been used to accelerate or approximate the solution of combinatorial and mixed-integer optimization problems~\citep{bengio2021}. Within the literature on machine learning for mixed-integer optimization, we distinguish three broad approaches. The first learns auxiliary information, such as active constraints or the values of binary variables, to reduce the size or complexity of the optimization problem. By predicting and fixing selected decision variables or eliminating inactive constraints, these methods can reduce the dimensionality and computational burden of the original problem~\citep{bertsimas2021,bertsimas2022}. However, incorrectly predicted variables or constraints may lead to a reduced problem for which neither feasibility nor optimality of the original problem is guaranteed.

A second approach uses machine learning to improve the performance of optimization solvers, for example by generating high-quality initial solutions as warm starts~\citep{Ahmed}, or by learning cutting planes and branching rules~\citep{alvarez,khalil,balcan,UCWarmStart}. Such methods can substantially accelerate mixed-integer optimization. However, an optimization problem still needs to be solved to obtain the final decision, which may remain computationally burdensome in real-time or high-frequency applications.

A third approach avoids solving an optimization problem at the time of decision making altogether by learning a direct mapping from observable features to decisions, commonly referred to as a policy~\citep{ban2019,liyanage2005,oroojlooyjadid,self,end}. Such policies can be represented by linear decision rules~\citep{ban2019,liyanage2005} or by more expressive function classes such as neural networks~\citep{oroojlooyjadid,self,end}. Once trained, the policy maps newly observed features directly to decisions without requiring a subsequent optimization problem to be solved. This makes direct policy learning particularly attractive in settings where decisions must be produced rapidly. However, when the learned decisions are used directly, feasibility with respect to the original optimization problem is not guaranteed.


Despite differences in architecture and application, much of the existing machine-learning literature for mixed-integer optimization, discussed above follows a predominantly \emph{predictive} paradigm. For each training instance, the underlying optimization problem is first solved offline to obtain an optimal binary decision, which is then used as a supervised-learning label. The learned model is therefore trained to reproduce instance-wise optimal decisions, without accounting for the objective values or feasibility of alternative decisions.

This limitation is particularly relevant to the third approach, where the learned policy is used directly at the time of decision making. If the chosen policy class cannot exactly reproduce the mapping from features to optimal binary decisions, minimizing prediction error over fixed labels need not yield the best policy within that class. A small prediction error may alter a binary decision and lead to infeasibility, while a slightly suboptimal but feasible decision may be better represented by the policy. This motivates a \emph{prescriptive} learning approach in which the feasible training decisions and policy parameters are optimized jointly, rather than treating instance-wise optimal decisions as fixed labels. However, even when prescriptive learning enforces feasibility across the training samples, feasibility is not guaranteed for unseen instances. The learned policy may still violate constraints because of finite-sample error or shifts in the underlying data distribution. We therefore ask: \textit{How can we learn direct binary decision policies that account for feasible alternatives during training while providing feasibility guarantees under distributional uncertainty?}

To address this question, our first contribution is a prescriptive learning framework for binary decision-making. The framework jointly optimizes the policy parameters and feasible decisions across the training scenarios using the original objective function and constraints. In contrast to predictive learning, the training decisions are endogenous to the learning problem, allowing the model to account for both their operational performance and their compatibility with the chosen policy class. Once trained, the policy maps newly observed features directly to binary decisions without requiring a new optimization problem to be solved at the time of decision making.
Our second contribution is a distributionally robust extension that strengthens the feasibility of the learned policy beyond the empirical training distribution. We formulate individual distributionally robust chance constraints using Wasserstein ambiguity sets centered at the empirical distribution. The Wasserstein radius and chance-constraint risk level provide tunable parameters that control the trade-off between operational performance and protection against sampling error and distributional shifts.

To the best of our knowledge, this is the first framework that combines prescriptive learning of direct linear policies for binary decisions with formal feasibility guarantees under distributional uncertainty. Closely related work on prescriptive decision trees for integer decisions~\citep{bertsimasprescriptive} considers contextual integer decision-making but does not provide distributionally robust feasibility guarantees of the form studied here. More broadly, established approaches to prescriptive learning and predict-then-optimize predominantly consider continuous decisions, continuous uncertain quantities, or optimization downstream of a prediction model~\citep{bertsimasprescriptive1,bertsimasholistic,smart}. Our framework instead focuses on direct binary decision policies and incorporates both the original feasibility structure and distributional robustness into the policy-training problem.

We demonstrate the proposed framework on two benchmark problems: unit commitment and staff allocation. Our out-of-sample experiments consider both settings in which the training and test distributions coincide and settings in which they differ. The results show that robustification may increase operational costs when the training and test distributions are well aligned, but can substantially reduce infeasibility costs under distributional shifts, resulting in lower total costs. These findings highlight the importance of accounting for both operational performance and feasibility when learning direct binary decision policies.

The remainder of the paper is structured as follows. Section~2 introduces the prescriptive learning framework and illustrates its distinction from classical predictive learning. Section~3 develops the distributionally robust chance-constrained formulation and its feasibility guarantees. Section~4 presents numerical results for the unit commitment and staff allocation case studies. Section~5 concludes.

\section{Prescriptive Learning}\label{sec:Method}
In this section, we develop a data-driven framework that couples classical machine learning with mixed-integer optimization to \emph{prescribe} decisions. We begin with preliminaries and assumptions, briefly review a predictive (supervised) baseline, and then introduce our prescriptive formulation.

\subsection{Preliminaries}

We consider mixed-binary linear programming problems with uncertainty in the parameters of the constraints. Let $\mathbf{x}\in\{0,1\}^N$ and $\mathbf{y}\in\mathbb{R}^K$ be vectors of binary and continuous decision variables, respectively, and the vectors $\mathbf{c}\in\mathbb{R}^N$ and $\mathbf{d}\in\mathbb{R}^K$ define the objective function. Let $\boldsymbol{\theta}:\Omega\rightarrow\mathbb{R}^{L_{\boldsymbol\theta}}$ be a random vector on some probability space $(\Omega, \mathbb{P})$, and let the matrices $\mathbf{A}(\boldsymbol{\theta})\in\mathbb{R}^{M\times N}$ and $\mathbf{E}(\boldsymbol{\theta})\in\mathbb{R}^{M\times K}$ and the vector $\mathbf{b}(\boldsymbol{\theta})\in\mathbb{R}^M$, all potentially depending on $\boldsymbol{\theta}$, define the linear constraints. Thus, the problem is as follows:
\begin{subequations}\label{eq:bin-base}
\begin{align}
\min_{\mathbf{x}\in\{0,1\}^N, \mathbf{y}\in\mathbb{R}^K}\ &\mathbf{c}^\top\mathbf{x}+\mathbf{d}^\top\mathbf{y}\\
\text{s.t.} \ &\mathbf{A}(\boldsymbol{\theta})\mathbf{x}+\mathbf{E}(\boldsymbol{\theta})\mathbf{y}\leq \mathbf{b}(\boldsymbol{\theta})
\end{align}
\end{subequations}


\paragraph{Example 1 (Staff allocation; pure binary).}
Let $x_n\in\{0,1\}$ indicate whether employee $n$ is assigned to a given shift for $n=1,\dots,N$. The problem is to minimize total staffing costs subject to coverage and workload constraints. We formulate it as a binary linear programming problem of the form \eqref{eq:bin-base} with $K=0$, and $\boldsymbol{\theta}$ encoding uncertain shift in demands that affects the coverage constraints. The specific formulation used in our experiments can be found in Section~4.

\paragraph{Example 2 (Unit commitment; mixed-binary).} In the so-called unit commitment problem, the binary variable $x_{jt}$ indicates whether power generator $j$ is committed (i.e., online/offline) at time $t$, whereas the continuous variable $y_{jt}$ represents the corresponding production level, for $j=1,\dots,J$ and $t=1,\dots,T$ (such that $N=K=TJ$). The objective accumulates operational costs, including fixed, start-up, and variable costs, while feasibility is governed by minimum uptime/downtime constraints, ramping restrictions, power balances, and transmission limits. The random vector $\boldsymbol{\theta}$ may encode uncertainty in demand and/or renewable supply. A detailed unit commitment formulation used in our experiments appears in Section~4.





\subsection{Assumptions} \label{assumptions}


\textcolor{black}{Let $\mathbf{z}\in \mathbb{R}^{L_{\mathbf{z}}}$ be a so-called feature vector representing contextual information available at the time of decision-making. We assume that the joint distribution of the random vectors $\mathbf{z}$ and $\boldsymbol{\theta}$ is unknown. However, the decision-maker has access to a dataset $\mathcal{D}_S=\{\mathbf{z}^s,\boldsymbol\theta^s\}_{s=1}^S$, where $\mathbf{z}^s$ is a sample from the feature vector $\mathbf{z}$ and $\boldsymbol\theta^s$ is a sample from the parameter vector $\boldsymbol\theta$ and $S$ is the number of samples.} The samples are assumed to be independently and identically distributed. Throughout the paper, we use the terms “features,” “contexts,” and “contextual information” interchangeably. 

\color{black}\begin{remark}
The feature vector $\mathbf{z}$ represents all observable information available at the time of decision-making, including contextual variables (e.g., time, location, past demand patterns), forecasts and/or summary statistics related to the uncertain parameters. Thus, $\mathbf{z}$ may include forecasts of $\boldsymbol{\theta}$, or summary statistics derived from historical realizations. In the numerical experiments of Section~\ref{results}, $\mathbf{z}$ contains forecasts of $\boldsymbol{\theta}$ that are available at the time of decision-making, whereas the realized values of $\boldsymbol{\theta}$ are revealed only after the decisions are made. In training, the realized values enter only through the constraints (and the labels of the classical approach); in testing, they are used to evaluate the decisions. In the illustrative examples of Subsections~\ref{example1} and~\ref{example2}, for simplicity of exposition, we consider the special case $\mathbf{z} = \boldsymbol{\theta}$. That is, we assume perfect information: $\boldsymbol{\theta}$ is known when the decision is made, which corresponds to a perfect forecast.
For further clarification, see the following examples.
\end{remark}\color{black}

\color{black}

\paragraph{Example 1 (Staff Allocation).}
In the staff allocation problem, the uncertain parameter $\boldsymbol{\theta}$ may represent the random number of customers arriving at a service center during a given shift, which determines the required number of staff to meet service-level constraints. The feature vector $\mathbf{z}$ could include calendar information (e.g., whether the day is a weekend or holiday), scheduled events, or historical average demand for that shift, information available before decisions are made. 

\paragraph{Example 2 (Unit Commitment).} 
In the unit commitment problem, the uncertain parameter $\boldsymbol{\theta}$ typically represents the random electricity demand at each bus or node in the grid during a given time period. These values directly affect the feasibility of power balance and transmission constraints but are unknown at the time commitment decisions are made. The contextual feature vector $\mathbf{z}$ may include time-of-day, day-of-week indicators, weather forecasts, or demand forecasts, all of which are observable in advance and can inform a policy.


The complexity of the mixed-binary linear programming problem is due to the binary decisions variables, and thus, these are our focus. Given a feature vector $\mathbf{z}$, our objective is to determine a `good' binary decision $\mathbf{x}$ conditional on this context. To that end, we aim to learn a linear policy $\boldsymbol{\pi}_{\mathbf{W}}: \mathbb{R}^{L_{\mathbf{z}}}\rightarrow \{0,1\}^N$ from a parametric hypothesis class $\Pi_{\mathcal{W}}:= \{\boldsymbol{\pi}_{\mathbf{W}}\}_{\mathbf{W} \in \mathcal{W}}$, which maps a feature vector $\mathbf{z}$ to a near-optimal binary decision $\mathbf{x}$ using the historical dataset $\mathcal{D}_S$. A parametric policy refers to a decision rule characterized by a finite-dimensional parameter matrix $\mathbf{W}$. We focus on linear parametric policies of the form, i.e., $\boldsymbol{\pi}_{\mathbf{W}}(\mathbf{z}) = \mathbf{W} \mathbf{z}$, where $\mathbf{W}\in\mathbb{R}^{N\times L_{\mathbf{z}}}$. \color{black} Note that the feature vector $\mathbf{z}$ may include a constant entry equal to one to account for a vector of intercepts of the linear policy, i.e., $\boldsymbol{\pi}_{\mathbf{W}}(\mathbf{z}) = \sum_{n=1}^{L_z-1}\mathbf{w}_nz_n+\mathbf{w}_{L_z}$, where $\mathbf{w}_n$ is column $n$ of $\mathbf{W}$. \color{black} The linear policies strike a balance between interpretability, tractability, and generalization, and enable efficient training using convex optimization techniques. For $n=1,\dots,N$, the mapping determines a binary decision $x_n$ such that $x_n=1$ if $\boldsymbol{\pi}_{\mathbf{W},n}(\mathbf{z})=(\mathbf{w}^n)^\top\mathbf{z}> 0.5$ and $x_n=0$ otherwise, where $\mathbf{w}^n$ is row $n$ of the matrix $\mathbf{W}$. This structure implicitly defines a rounding behavior, values above 0.5 are interpreted as 1, and those below as 0, allowing the policy to be trained using continuous optimization techniques while maintaining a binary interpretation of decisions.

\subsection{Classical Predictive Learning}
\textcolor{black}{One way to determine a policy $\boldsymbol\pi_\mathbf{W}$ is to use classical predictive learning. For each sample from the dataset $\mathcal{D}_S$, we solve the optimization problem~\eqref{eq:bin-base} with $\boldsymbol{\theta}^s$ and obtain an optimal binary decision $\mathbf{x}^{*,s}$.} This gives us a set  of supervised training data $\{(\mathbf{z}^s,\mathbf{x}^{*,s})\}_{s=1}^S$. The predictive learning problem minimizes the average distance between $\boldsymbol\pi_\mathbf{W}(\mathbf{z}^s)$ and $\mathbf{x}^{*,s}$ across all samples:
\begin{subequations}\label{classical}
\begin{align} 
    \min_{\mathbf{W}\in \mathcal{W}} \frac{1}{S}&\sum_{s=1}^S||\boldsymbol\pi_\mathbf{W}(\mathbf{z}^s)-\mathbf{x}^{*,s}||
\end{align}
\end{subequations}
where the operator $||.||$ denotes a suitable norm (e.g., L1 or L2). 

The disadvantages of the classical predictive approaches are as follows.
    The supervised learning paradigm requires labeled data. In the context of decision-making, this means that for every sample of the feature vector $\textbf{z}^s$, the optimal binary decision $\mathbf{x}^{*,s}$ must be obtained by solving the underlying optimization problem for $\boldsymbol{\theta}^s$. This process, which is referred to as the data generating process is often computationally expensive and time-consuming, especially in large-scale or complex settings. Furthermore, the optimization problem may admit multiple optimal solutions or exhibit symmetries, making it challenging to generate consistent and informative training labels. As a result, careful orchestration of the data generation process is necessary to ensure the effectiveness and reliability of the supervised learning pipeline.
Furthermore, a key limitation of predictive learning is its inability to account for counterfactuals, the outcomes of decisions that were not actually made. Consequently, the learned model only captures observed optimal decisions, lacking insights into the performance of alternative decisions. This shortcoming becomes particularly critical in constrained optimization problems, where the learned optimal decisions often lie near the boundary of the feasible set. Such solutions are vulnerable to even small prediction errors, which can render them infeasible. 

To address these issues, we propose a prescriptive learning approach that bypasses the data generation process by training a
machine learning model with a loss function that captures the original objective function and constraints of the problem.  The details of this formulation are presented in the following subsections.

   
\subsection{Proposed Prescriptive Formulation}


Although there exists a $\mathbf{y}^s\in\mathbb{R}^K$ such that $\mathbf{A}(\boldsymbol{\theta}^{s})\mathbf{x}^{*,s}+\mathbf{E}(\boldsymbol{\theta}^{s})\mathbf{y}^{s}\leq\mathbf{b}(\boldsymbol{\theta}^s)$ for all $s=1,\dots,S$, there may not exist an $\mathbf{y}^s\in\mathbb{R}^K$ such that $\mathbf{A}(\boldsymbol{\theta^{s}})\mathbf{x}^{s}+\mathbf{E}(\boldsymbol{\theta}^{s})\mathbf{y}^{s}\leq\mathbf{b}(\boldsymbol{\theta}^{s})$ for all $s=1,\dots,S$, where $x_n^s=1$ if $\boldsymbol\pi_{\mathbf{W},n}(\mathbf{z}^{s})>0.5$ and $x_n^s=0$ otherwise, for $n=1,\dots,N$. That is, although the optimal decisions from the data generating process are feasible, predicted decisions may not be. To enhance the feasibility of the policy output, we begin by augmenting the classical supervised learning formulation~\eqref{classical} with the original problem constraints. This leads to the following intermediate formulation:

\begin{subequations}\label{classical_feasibility}
\begin{align} 
    \min_{\begin{subarray}{l} 
      \mathbf{x}^{s}\in \{0,1\}^N  ,\\
      \mathbf{y}^{s}\in \mathbb{R}^K,\mathbf{W}\in \mathcal{W}
    \end{subarray}}
    \frac{1}{S}&\sum_{s=1}^S \big[ \,||\boldsymbol\pi_\mathbf{W}(\mathbf{z}^{s})-\mathbf{x}^{*,s}|| + 
    ||\boldsymbol\pi_\mathbf{W}(\mathbf{z}^{s})-\mathbf{x}^{s}||\,\big] \\
     \text{s.t.} \quad & \boldsymbol\pi_{\mathbf{W},n}(\mathbf{z}^{s}) \leq 0.5 + x^{s}_n , \enspace n=1,\dots,N, s=1,\dots,S, \label{classical_feasibility_con1}\\
     & \boldsymbol\pi_{\mathbf{W},n}(\mathbf{z}^{s}) \geq x^{s}_n-0.5,\enspace n=1,\dots,N,s=1,\dots,S , \label{classical_feasibility_con2}\\
     & {\mathbf{a}^m(\boldsymbol{\theta}^s)}^{\top}\mathbf{x}^{s}+{\mathbf{e}^m(\boldsymbol{\theta}^s)}^{\top}\mathbf{y}^{s}\leq b_m (\boldsymbol{\theta}^s) , \enspace m=1,\dots,M, s=1,\dots,S. \label{classical_feasibility_con3}
\end{align}
\end{subequations}

Note that the vector $\mathbf{a}^m(\boldsymbol{\theta}^s)$ and $\mathbf{e}^m(\boldsymbol{\theta}^s)$ denote the row $m$ of $\mathbf{A}(\boldsymbol{\theta}^s)$ and $\mathbf{E}(\boldsymbol{\theta}^s)$, respectively. The formulation jointly determines the binary decisions $\mathbf{x}^{s}$, continuous decisions $\mathbf{y}^{s}$, and policy parameters $\mathbf{W}$ to ensure feasibility. Constraints~\eqref{classical_feasibility_con1}–\eqref{classical_feasibility_con2} effectively convert the continuous output of the policy $\boldsymbol\pi_\mathbf{W}(\mathbf{z}^{s})$ into binary decisions by enforcing a threshold at 0.5. Specifically, if $\boldsymbol\pi_{\mathbf{W},n}(\mathbf{z}^{s}) > 0.5$, then ${x}_n^{s} = 1$; otherwise, ${x}_n^{s} = 0$. Note that the objective includes the distance term $\|\boldsymbol\pi_\mathbf{W}(\mathbf{z}^{s}) - \mathbf{x}^{s}\|$ to avoid a trivial solution with $x_n^s=0.5, n=1,\dots,N$ for all $s=1,\dots,S$ that would minimize the integer-relaxed and unconstrained version of the problem. Constraint~\eqref{classical_feasibility_con3} ensures that the decisions satisfy the problem constraints. Thus, compared to~\eqref{classical}, this formulation not only learns a policy that predicts decisions close to the ground-truth optimal solutions, but also ensures feasibility in the original problem.  

A remaining limitation is that the objective in~\eqref{classical_feasibility} treats all deviations of the policy output from the ground-truth optimal solutions $\mathbf{x}^{*,s}$ equally, regardless of their impact on the actual cost. In practice, suboptimality may vary significantly across instances, and small deviations in some cases may incur disproportionately higher costs. To address this issue, we propose a prescriptive learning formulation that directly minimizes the original objective function, rather than relying on the distance between the policy output and an optimal solution. At the same time, we retain the distance term $\|\boldsymbol\pi_\mathbf{W}(\mathbf{z}^{s}) - \mathbf{x}^{s}\|$ in the objective. The resulting proposed prescriptive formulation is:
 
\begin{subequations}\label{prescriptive}
\begin{align} 
    \min_{\begin{subarray}{l} 
      \mathbf{x}^{s}\in \{0,1\}^N  ,\\
      \mathbf{y}^{s}\in \mathbb{R}^K,\mathbf{W}\in \mathcal{W}
    \end{subarray}}
    \frac{1}{S}&\sum_{s=1}^S \big[\,\mathbf{c}^\top\mathbf{x}^{s}+\mathbf{d}^\top\mathbf{y}^{s}+||\boldsymbol\pi_\mathbf{W}(\mathbf{z}^{s})-\mathbf{x}^{s}||\,\big] \label{prescriptive_obj} \\
     \text{s.t.} \quad & \text{(\ref{classical_feasibility_con1})-(\ref{classical_feasibility_con3})}.
\end{align}
\end{subequations}

In this formulation, the binary decisions $\mathbf{x}^{s}$, continuous decisions $\mathbf{y}^{s}$, and policy parameters $\mathbf{W}$ are jointly optimized to minimize the costs while ensuring feasibility. The proposed approach adopts prescriptive learning, meaning it does not rely on the dataset $\mathcal{D}_S$ of precomputed optimal decisions \(\mathbf{x}^{*,s}\). Instead of mimicking these idealized labels, our method directly optimizes the policy’s decisions for feasibility and performance in the original problem. The key advantage of the proposed training problem~\eqref{prescriptive} is its explicit handling of the \textit{bias error} arising from restricting the policy to a parametric hypothesis class $\Pi_{\mathcal{W}}$ (e.g., linear policies). Unlike the classical approach~\eqref{classical}, which trains policies to reproduce the ground-truth decisions \(\mathbf{x}^{*,s}\) while ignoring the cost of deviating from them, our prescriptive formulation jointly i) accounts for the quality of \(\boldsymbol{\pi}_{\mathbf{W}}\) as measured by the original objective, and ii) enforces feasibility by prioritizing constraint satisfaction over label fidelity. By contrast,~\eqref{classical}  fails to ensure that the learned policy \(\boldsymbol{\pi}_\mathbf{W}\) produces decisions that are \emph{actionable} (i.e., satisfy constraints) or \emph{prescriptively optimal} within \(\Pi_{\mathcal{W}}\).

\subsection{Linear Policy for Binary Decisions}
In this paper, we focus on linear policies because they are simple, computationally efficient, and interpretable to end users. 
Therefore, we can re-write problem (\ref{prescriptive}) as follows:
\begin{subequations}\label{prescriptive_linear}
\begin{align}     \min_{\begin{subarray}{l} 
      \mathbf{x}^s\in \{0,1\}^N,\mathbf{y}^s\in \mathbb{R}^K,\\
      \quad\mathbf{W}\in \mathcal{W}
    \end{subarray}}
    \frac{1}{S}&\sum_{s=1}^S \big[ \,\mathbf{c}^\top\mathbf{x}^{s}+\mathbf{d}^\top\mathbf{y}^{s}+||\mathbf{W} \mathbf{z}^{s}-\mathbf{x}^{s}||\,\big] \label{prescriptive_objectivefunction}\\
     \text{s.t.} \quad  
     & ({\mathbf{w}^n})^\top \mathbf{z}^{s} \leq 0.5 + x^{s}_n,\enspace n=1,\dots,N, s=1,\dots,S,   \label{prescrptive_linear_con1}\\
     & ({\mathbf{w}^n})^\top \mathbf{z}^{s} \geq x^{s}_n-0.5,\enspace n=1,\dots,N, s=1,\dots,S,    \label{prescrptive_linear_con2}\\  &\text{(\ref{classical_feasibility_con3})}\nonumber.
\end{align}
\end{subequations}

\subsection{Illustrative Example} \label{example1}
To motivate our proposed prescriptive learning approach, we provide here a simple yet illustrative example where classical supervised learning fails to produce feasible decisions, while our method avoids such issues by directly incorporating the constraints  in the training process. 

Consider a problem with a single source of uncertainty $\theta\in\mathbb{R}$ and a binary decision variable $x\in\{0,1\}$. Assume that $\mathbf{z}=(\theta,1)$ such that \( \pi(\mathbf{z})=w_1\theta+w_0\) is a linear function of 
\(\theta\), augmented with an intercept term. We assume perfect information, i.e., the realized value of $\theta$ is used as the feature ($z=\theta$). This is the most favorable setting for the classical approach, since the feature contains all information needed to determine the optimal decision. Any infeasibility of the classical policy can therefore be attributed to learning from labels rather than to forecast errors. In practice, $\mathbf{z}$ would represent a forecast of $\theta$ available before the decision (see Remark~1). The problem is to determine $x$ such that the objective function is minimized, subject to the constraints:
\begin{subequations}
\begin{align}
\min_{x, b_1, b_2, b_3} \quad & -x \\
\text{s.t.} \quad 
& \theta x - 0.8 \le 0 \label{eq:constr1} \\
& \theta \le 0.4 + M(1 - b_1) \label{eq:constr2} \\
& \theta \ge 0.4 - M(1 - b_2) \label{eq:constr3} \\
& \theta \le 0.7 + M(1 - b_2) \label{eq:constr4} \\
& \theta \ge 0.7 - M(1 - b_3) \label{eq:constr5} \\
& x \le 1 - b_2 \label{eq:constr6} \\
& b_1 + b_2 + b_3 = 1 \label{eq:constr7} \\
& x \in \{0,1\}, \quad b_1, b_2, b_3 \in \{0,1\} \label{eq:constr8}
\end{align}
\end{subequations}
where $b_1,b_2,$ and $b_3$ are auxiliary binary variables and $M$ is a sufficiently large constant. Note that the objective \( -x \) encourages the decision-maker to choose \( x^* = 1 \) when it is feasible. The constraint (\ref{eq:constr1}) ensures \( x^*=0 \) 
for $\theta> 0.8$. 
Constraints (\ref{eq:constr2})-(\ref{eq:constr7}) partition the space of $\theta$ into three disjoint regions $\theta\le 0.4$, $0.4<\theta<0.7$ and $\theta\geq 0.7$. For $0.4<\theta<0.7$ (and \(b_1=b_3=0, b_2 = 1 \)), \( x^*=0 \). 

Let us consider the following set of $S=14$ sample values:
\[
\{\theta^s\}_{s=1}^S=\{0.1, 0.2, 0.3, 0.4, 0.5, 0.6, 0.7, 0.8, 0.9, 1.0, 1.1, 1.2, 1.3, 1.4\}.
\]
For each \( \theta^s\), we solve the optimization problem and obtain the corresponding optimal decision \( x^{*,s} \). The optimal decisions are:
\[
\{x^{*,s}\}_{s=1}^S=\{1, 1, 1, 1, 0, 0, 1, 1, 0, 0, 0, 0, 0, 0\}.
\]

A standard supervised learning approach fits a regression model \( \pi(z) \approx x^* \), and makes decisions by thresholding, i.e., if \( \pi(z) > 0.5 \), then \( x^* = 1 \); otherwise \( x^* = 0 \). Therefore, this model is unaware of the underlying feasibility logic encoded in the optimization formulation. 

\begin{figure}[!htbp]
    \centering
    \includegraphics[width=0.8\linewidth]{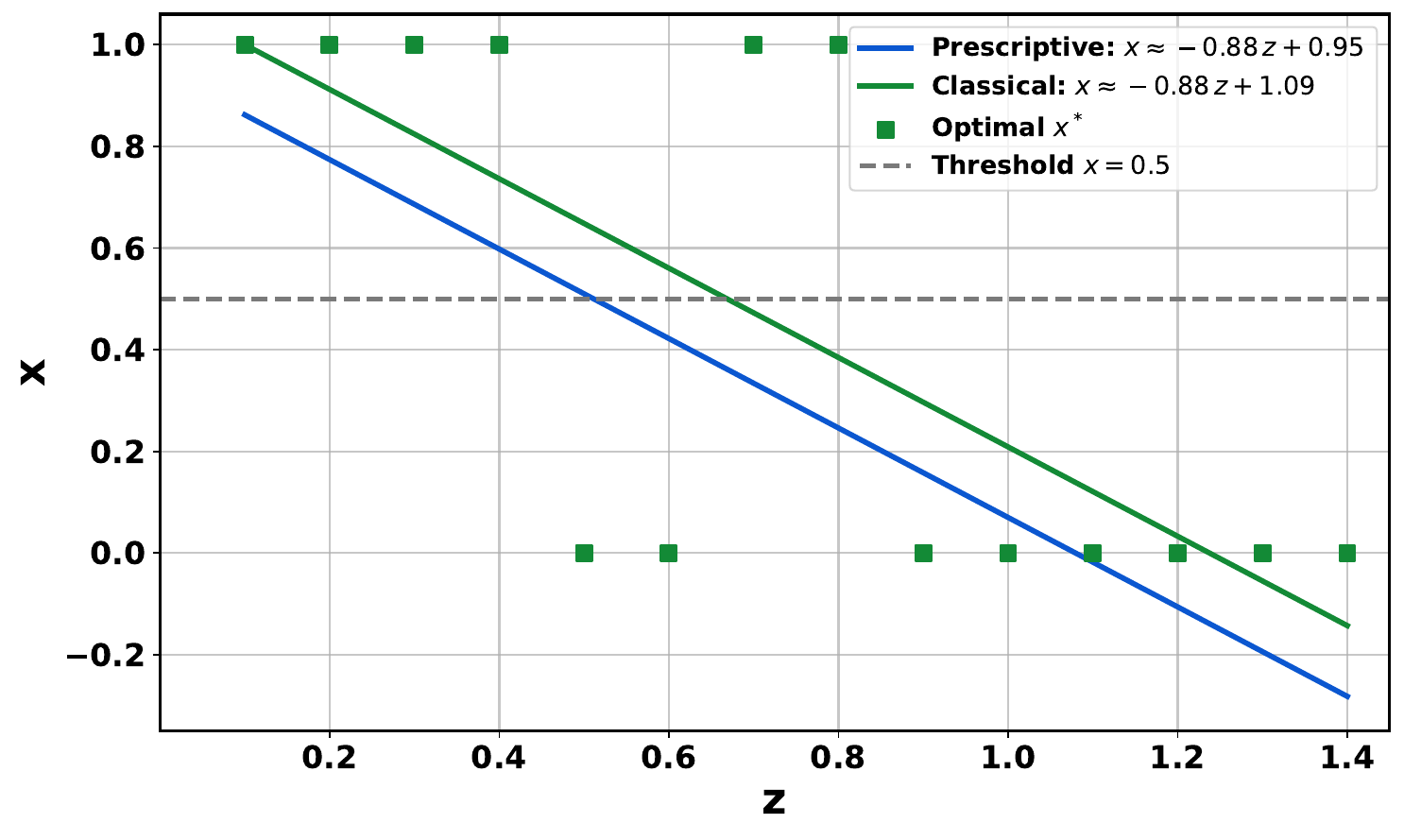}
    \caption{Comparison of the classical and prescriptive linear policies for the illustrative example. The square markers represent the optimal binary decisions for the training samples, and the dashed horizontal line denotes the decision threshold at \(0.5\). The classical policy is fitted solely to the optimal labels and may therefore prescribe infeasible decisions, whereas the prescriptive policy incorporates the original optimization constraints during training and preserves feasibility at the expense of some loss in optimality.}
    \label{fig:regression_comparison}
\end{figure}

Figure~\ref{fig:regression_comparison} illustrates the comparison between classical supervised learning (green line) and the proposed prescriptive method (blue line). As shown in Figure~\ref{fig:regression_comparison}, the classical regression model predicts values \( \pi(z) > 0.5 \) in regions where the feasible solution requires \( x = 0 \).  For example,
for \( \theta = 0.5\) and \( \theta = 0.6\),  \( x^* = 0 \). However, the regression model may predict \( \pi(0.5,1) > 0.5 \), leading to an infeasible solution. These violations occur because the model is trained solely to minimize the prediction error between \( \pi(z) \) and \( x^* \), without incorporating the feasibility constraints. In contrast, prescriptive learning guarantees that all predicted decisions are feasible (blue line). However, it may still produce suboptimal results, for instance, it predicts \( x^* = 0 \) for \( \theta = 0.7 \) and \( \theta = 0.8 \), even though the optimal decision in both cases is \( x^* = 1 \). Despite this, the predictions remain feasible under the given constraints.

\section{Distributionally Robust Prescriptive Learning} \label{section3}
The purpose of the prescriptive learning problem \eqref{prescriptive_linear} is to learn decisions from a finite set of training samples that perform well on unseen instances. This objective implicitly relies on two assumptions: first, that the joint distribution of $(\mathbf{z},\boldsymbol{\theta})$ during training coincides with the distribution encountered at testing; and second, that the empirical distribution induced by the samples $\{(\mathbf{z}^s,\boldsymbol{\theta}^s)\}_{s=1}^S$ is a sufficiently accurate approximation of that distribution. Both assumptions may fail in practice because of limited sample availability, sampling error, or nonstationarity in the data-generating process \citep{MohajerinEsfahani2018,gao2023}. Consequently, a policy that is feasible on the training samples may fail to remain feasible out of sample. This issue is especially important in applications where infeasible decisions lead to high operational or safety-related costs. We therefore focus on robustifying the \emph{feasibility} of the learned policy, rather than its objective value.
A classical way to protect feasibility against uncertainty is to adopt a robust optimization perspective, requiring the policy-consistency constraints \eqref{prescrptive_linear_con1}--\eqref{prescrptive_linear_con2} and the original feasibility constraints \eqref{classical_feasibility_con3} to hold for every realization of $(\mathbf{z},\boldsymbol{\theta})$ in a prescribed uncertainty set $\mathcal{U}_{\mathbf{z},\boldsymbol{\theta}}$. To express this idea abstractly, let $\mathbf{x}(\mathbf{z},\boldsymbol{\theta})$ and $\mathbf{y}(\mathbf{z},\boldsymbol{\theta})$ denote realization-dependent decisions associated with a given realization of the exogenous uncertainty. The robust counterpart can then be written as
\begin{subequations}\label{robust}
\begin{align}
&({\mathbf{w}^n})^\top \mathbf{z} \le 0.5 + x_n(\mathbf{z},\boldsymbol{\theta}),
\enspace n=1,\dots,N,\ \forall (\mathbf{z},\boldsymbol{\theta}) \in \mathcal{U}_{\mathbf{z},\boldsymbol{\theta}}, \\
&({\mathbf{w}^n})^\top \mathbf{z} \ge x_n(\mathbf{z},\boldsymbol{\theta}) - 0.5,
\enspace n=1,\dots,N,\ \forall (\mathbf{z},\boldsymbol{\theta}) \in \mathcal{U}_{\mathbf{z},\boldsymbol{\theta}}, \\
&{\mathbf{a}^m(\boldsymbol{\theta})}^{\top}\mathbf{x}(\mathbf{z},\boldsymbol{\theta})
+{\mathbf{e}^m(\boldsymbol{\theta})}^{\top}\mathbf{y}(\mathbf{z},\boldsymbol{\theta})
\le b_m(\boldsymbol{\theta}),
\enspace m=1,\dots,M,\ \forall (\mathbf{z},\boldsymbol{\theta}) \in \mathcal{U}_{\mathbf{z},\boldsymbol{\theta}}.
\end{align}
\end{subequations}
Although formulation \eqref{robust} provides strong feasibility protection, it may be overly conservative for two reasons. First, it treats all realizations in $\mathcal{U}_{\mathbf{z},\boldsymbol{\theta}}$ as equally relevant. Second, it imposes hard feasibility for every realization, allowing no violations whatsoever. To reduce this conservatism, one may instead adopt a chance-constrained framework, which permits a small probability of violation under a reference probability distribution. If $\hat P_{\mathbf{z},\boldsymbol{\theta}}$ denotes the empirical distribution of $(\mathbf{z},\boldsymbol{\theta})$, then the corresponding individual chance constraints are
\begin{subequations}\label{empirical_chance}
\begin{align}
&\mathbb{P}^{\hat P_{\mathbf{z},\boldsymbol{\theta}}}
\big[({\mathbf{w}^n})^\top \mathbf{z} \le 0.5 + x_n(\mathbf{z},\boldsymbol{\theta})\big]
\ge 1-\epsilon,
\enspace n=1,\dots,N, \\
&\mathbb{P}^{\hat P_{\mathbf{z},\boldsymbol{\theta}}}
\big[({\mathbf{w}^n})^\top \mathbf{z} \ge x_n(\mathbf{z},\boldsymbol{\theta}) - 0.5\big]
\ge 1-\epsilon,
\enspace n=1,\dots,N, \\
&\mathbb{P}^{\hat P_{\mathbf{z},\boldsymbol{\theta}}}
\big[
{\mathbf{a}^m(\boldsymbol{\theta})}^{\top}\mathbf{x}(\mathbf{z},\boldsymbol{\theta})
+
{\mathbf{e}^m(\boldsymbol{\theta})}^{\top}\mathbf{y}(\mathbf{z},\boldsymbol{\theta})
\le b_m(\boldsymbol{\theta})
\big]
\ge 1-\epsilon,
\enspace m=1,\dots,M,
\end{align}
\end{subequations}
where $\epsilon \in [0,1]$ is a risk parameter controlling the allowed probability of constraint violation.
However, the empirical distribution $\hat P_{\mathbf{z},\boldsymbol{\theta}}$ may itself be a poor approximation of the true distribution and may fail to capture distributional shifts at deployment. To address this limitation, we adopt a distributionally robust chance-constrained formulation. Distributionally robust chance-constrained optimization seeks decisions that satisfy the chance constraints under the worst-case probability distribution within a prescribed ambiguity set constructed from the available data. Equivalently, the chance constraints are required to hold for all distributions in the ambiguity set.
In this work, we focus on distributionally robust chance constraints under Wasserstein ambiguity sets \citep{MohajerinEsfahani2018,chen2022} and retain the objective function of the prescriptive model introduced in the previous section, namely \eqref{prescriptive_obj}. Throughout this section, we assume that the uncertainty affects only the right-hand side of the constraints, i.e., $\mathbf{A}(\boldsymbol{\theta}) \equiv \mathbf{A}$ and $\mathbf{E}(\boldsymbol{\theta}) \equiv \mathbf{E}$, and that $\mathbf{b}(\boldsymbol{\theta})$ is affine in $\boldsymbol{\theta}$. This assumption is needed because the reformulation of \citet{chen2022} used in this section requires each constraint to be affine in the random vector. If the coefficients of $\mathbf{x}$ and $\mathbf{y}$ depended on $\boldsymbol{\theta}$, the constraints would contain products of $\boldsymbol{\theta}$ with $\mathbf{x}$ and $\mathbf{y}$, and this structure would be lost. The prescriptive formulation of Section~\ref{sec:Method} does not require this assumption.

To preserve computational tractability, we restrict attention to individual chance constraints and consider the following distributionally robust counterparts:
\begin{subequations}\label{dro_chance}
\begin{align}
&\inf_{Q\in\mathcal{P}_{\mathbf{z},\boldsymbol{\theta}}}
\mathbb{P}^{Q}\big[({\mathbf{w}^n})^\top \mathbf{z} \le 0.5 + x_n(\mathbf{z},\boldsymbol{\theta})\big]
\ge 1-\epsilon,
\enspace n=1,\dots,N, \label{DRO_constraint1}\\
&\inf_{Q\in\mathcal{P}_{\mathbf{z},\boldsymbol{\theta}}}
\mathbb{P}^{Q}\big[({\mathbf{w}^n})^\top \mathbf{z} \ge x_n(\mathbf{z},\boldsymbol{\theta}) - 0.5\big]
\ge 1-\epsilon,
\enspace n=1,\dots,N, \label{DRO_constraint2}\\
&\inf_{Q\in\mathcal{P}_{\mathbf{z},\boldsymbol{\theta}}}
\mathbb{P}^{Q}\big[
{\mathbf{a}^m}^{\top}\mathbf{x}(\mathbf{z},\boldsymbol{\theta})
+
{\mathbf{e}^m}^{\top}\mathbf{y}(\mathbf{z},\boldsymbol{\theta})
\le b_m(\boldsymbol{\theta})
\big]
\ge 1-\epsilon,
\enspace m=1,\dots,M. \label{DRO_constraint3}
\end{align}
\end{subequations}
Here, $\mathbb{P}^{Q}[\cdot]$ denotes probability under the distribution $Q$, and $\mathcal{P}_{\mathbf{z},\boldsymbol{\theta}}$ is an ambiguity set of probability distributions for $(\mathbf{z},\boldsymbol{\theta})$. Thus, \eqref{dro_chance} requires each constraint to be satisfied with probability at least $1-\epsilon$ under the worst-case distribution in $\mathcal{P}_{\mathbf{z},\boldsymbol{\theta}}$. Under individual chance constraints, the worst-case distribution may differ from one constraint to another. For simplicity, we use the same risk parameter $\epsilon$ for all constraints, although the framework could be extended to allow constraint-specific risk levels. If 
$\mathbf{z}$ contains a perfect forecast of 
$\boldsymbol{\theta}$ (i.e., 
$\mathbf{z}^s=\boldsymbol{\theta}^s$
 for all training samples), then the ambiguity set and decision functions depend only on 
$\mathbf{z}$. We next define the ambiguity set $\mathcal{P}_{\mathbf{z},\boldsymbol{\theta}}$.
\subsection{Wasserstein Ambiguity Set}
To account for sampling error and possible distributional shifts between training and testing, we construct a Wasserstein ambiguity set centered at the empirical distribution $\widehat P$. Following \citep{chen2022}, we collect the exogenous random quantities in the vector
$\boldsymbol{\xi}:=(\mathbf{z},\boldsymbol{\theta}) \in \mathbb{R}^{L},$ where $L=L_{\mathbf z}+L_{\boldsymbol{\theta}},$
and denote its training realizations by
\[
{\boldsymbol{\xi}}^{\,s}
:=
(\mathbf{z}^s,\boldsymbol{\theta}^s),
\qquad s=1,\dots,S.
\]
The empirical distribution supported on the training samples is then
\[
\widehat P
=
\frac{1}{S}\sum_{s=1}^S \delta_{{\boldsymbol{\xi}}^{\,s}},
\]
where $\delta_{{\boldsymbol{\xi}}^{\,s}}$ denotes the Dirac measure at ${\boldsymbol{\xi}}^{\,s}$.
Let $\|\cdot\|$ be a ground norm on $\mathbb{R}^{L}$. The type-1 Wasserstein distance between two probability distributions $Q_1$ and $Q_2$ on $\mathbb{R}^{L}$ is defined as
\begin{equation}
d_W(Q_1,Q_2)
=
\inf_{\Pi\in\Gamma(Q_1,Q_2)}
\int_{\mathbb{R}^{L}\times\mathbb{R}^{L}}
\|\boldsymbol{\xi}_1-\boldsymbol{\xi}_2\|
\, d\Pi(\boldsymbol{\xi}_1,\boldsymbol{\xi}_2),
\end{equation}
where $\Gamma(Q_1,Q_2)$ denotes the set of all couplings of $Q_1$ and $Q_2$, that is, all joint distributions with marginals $Q_1$ and $Q_2$. For a prescribed radius $\rho\ge 0$, we define the Wasserstein ambiguity set as
\begin{equation}
\mathcal{F}(\rho)
=
\left\{
Q\in\mathcal{P}(\mathbb{R}^{L})
:
d_W(Q,\widehat P)\le \rho
\right\}.
\label{wasserstein_ball}
\end{equation}
The radius $\rho$ can be interpreted as a transportation budget: every distribution in $\mathcal{F}(\rho)$ can be obtained from the empirical distribution $\widehat P$ by redistributing probability mass at total transportation cost at most $\rho$. Hence, feasibility is required not only under the empirical distribution, but under all distributions that lie within the Wasserstein ambiguity set. In particular, the learned policy is protected against sampling error and moderate distributional shifts whenever the true test distribution belongs to $\mathcal{F}(\rho)$.
\subsection{Reformulation under Wasserstein Ambiguity Sets}
\label{subsec:augmented_reformulation}
We now derive a tractable reformulation of the individual distributionally
robust chance constraints in \eqref{dro_chance}. The reformulation of \citet{chen2022}, which we use below, applies
to chance constraints whose safety set is defined over the random vector
appearing in the Wasserstein ambiguity set. In \eqref{dro_chance}, however,
the feasibility conditions involve not only the exogenous random vector
$\boldsymbol{\xi}:=(\mathbf{z},\boldsymbol{\theta})$, but also the decisions
$\mathbf{x}$ and $\mathbf{y}$, which may depend on the realization of
$\boldsymbol{\xi}$. Hence, the safety conditions cannot be written solely as
functions of the exogenous uncertainty unless the decision mappings
$\mathbf{x}(\mathbf{z},\boldsymbol{\theta})$ and
$\mathbf{y}(\mathbf{z},\boldsymbol{\theta})$ are specified explicitly.
A common approach in multistage or adjustable optimization is to approximate
continuous recourse decisions using affine decision
rules~\citep{ben-tal2004,kuhn2011}. When recourse decisions include integer
variables, another common strategy is to restrict attention to a finite set of
candidate decision rules~\citep{bertsimas2010finitead}. While these
approximations can improve tractability, they are not fully aligned with our
main objective of preserving feasibility. In particular, restricting continuous
recourse decisions to a prescribed functional class, or restricting binary
recourse decisions to a finite set of candidate policies, may eliminate
feasible reactions that would otherwise be available under new realizations of
$(\mathbf{z},\boldsymbol{\theta})$. This reduction of the recourse feasible set
can lead to excessive conservativeness and may even prevent the identification
of any feasible policy. Although our distributionally robust
chance-constrained formulation is itself conservative, its conservativeness is
controlled explicitly through the hyperparameters $\epsilon$ and $\rho$, which
allow the decision-maker to tune the trade-off between robustness and
performance. By contrast, conservativeness induced by restricting the feasible
set of recourse actions is structural; there is no mechanism for adjusting it
when the resulting problem becomes infeasible.
\color{black}To avoid this limitation, we adopt a different strategy. Recall that
$\boldsymbol{\xi}:=(\mathbf{z},\boldsymbol{\theta})$ denotes the primitive
exogenous random vector. For any fixed policy and recourse mechanism, there
exists a mapping $H:\boldsymbol{\xi}\mapsto(\mathbf{x},\mathbf{y})$ such that
$(\mathbf{x},\mathbf{y})=H(\mathbf{z},\boldsymbol{\theta})$. Hence,
$\boldsymbol{\omega}:=(\mathbf{z},\boldsymbol{\theta},\mathbf{x},\mathbf{y})
=(\boldsymbol{\xi},H(\boldsymbol{\xi}))$ is a legitimate random vector whose
true distribution is the pushforward
$P_{\boldsymbol{\omega}}=(\mathrm{Id},H)_\#P_{\boldsymbol{\xi}}$, concentrated
on the graph
$\mathcal{G}_H=\{(\boldsymbol{\xi},H(\boldsymbol{\xi})):\boldsymbol{\xi}\in
\mathcal{Z}\times\Theta\}$. The mapping $H$ has no closed form, however:
evaluating it at an unseen realization $(\mathbf{z},\boldsymbol{\theta})$ is
equivalent to solving the MIP~\eqref{eq:bin-base}, which is precisely what we
seek to avoid at deployment. Rather than characterizing $H$ explicitly, we
work directly in the lifted space
$\mathcal{Z}\times\Theta\times\mathcal{X}\times\mathcal{Y}$ and construct a
Wasserstein ambiguity set around the \emph{lifted training samples}
\[
\boldsymbol{\omega}^s
:=(\mathbf{z}^s,\boldsymbol{\theta}^s,\mathbf{x}^s,\mathbf{y}^s),
\quad s=1,\dots,S,
\]
where $(\mathbf{x}^s,\mathbf{y}^s)$ are the feasible training decisions
determined jointly with the policy parameters $\mathbf{W}$ in the prescriptive
training problem~\eqref{prescriptive_linear}. \color{black} These are endogenous marks
attached to each exogenous observation during training; they are not
historically observed random variables. The empirical measure
\[
\widehat{P}_{\boldsymbol{\omega}}
=\frac{1}{S}\sum_{s=1}^S\delta_{\boldsymbol{\omega}^s}
\]
is a decision-dependent empirical distribution in the lifted space, and can be
interpreted as an approximation of the true induced joint distribution
$P_{\boldsymbol{\omega}}$.
For numerical stability, we normalize the exogenous random variables
$\mathbf{z}$ and $\boldsymbol{\theta}$, as well as the continuous recourse
variable $\mathbf{y}$. The constraints~\eqref{DRO_constraint3} are then
written with respect to these normalized variables. No normalization is required for the binary variable $\mathbf{x}$, since $\mathbf{x} \in \{0,1\}^N$. However, inside the ambiguity set we relax $\mathbf{x}$ to $[0,1]^N$. This relaxation is required to apply the reformulation of \citet{chen2022} used below, which assumes that the random vector takes values in a continuous space. If $\mathbf{x}$ were kept binary inside the ambiguity set, the closed-form expression for the distance between each training sample and the constraint-violation region would no longer apply. Note that the relaxation applies only inside the ambiguity set, and the training decisions $\mathbf{x}^s$ remain binary in the optimization model. We define the
Wasserstein ambiguity set as
\begin{align}
\mathcal{P}_{\boldsymbol{\omega}}(\rho)
=
\left\{
Q\in\mathcal{P}_0\!\left(
[0,1]^{L_{\mathbf{z}}}
\times[0,1]^{L_{\boldsymbol{\theta}}}
\times[0,1]^N
\times[0,1]^K
\right)
:W(Q,\widehat{P}_{\boldsymbol{\omega}})\le\rho
\right\}.
\label{eq:ambiguity_omega}
\end{align}
The ambiguity set in~\eqref{eq:ambiguity_omega} is a conservative data-driven
approximation, as it does not enforce the functional dependence of
$(\mathbf{x},\mathbf{y})$ on $(\mathbf{z},\boldsymbol{\theta})$, and it allows
$\mathbf{x}$ to take fractional values inside the ball, further enlarging the
set of adversarial distributions. In practice, the degree of conservativeness
is controlled through the radius $\rho$ and the risk level $\epsilon$, which
are treated as tunable hyperparameters.
Using this augmented ambiguity set, each individual distributionally robust
chance constraint in~\eqref{dro_chance} can be written compactly as
\begin{align}
\inf_{Q\in\mathcal{P}_{\boldsymbol{\omega}}(\rho)}
\mathbb{P}^Q\!\left[
\mathbf{a}^i(\mathbf{W})^{\top}\boldsymbol{\omega}+\mathbf{b}^i\le 0
\right]
\ge 1-\epsilon,
\qquad i=1,\dots,I,
\label{DRCC_compact}
\end{align}
where $I$ denotes the number of individual distributionally robust chance
constraints. Here, $\boldsymbol{\omega} = (\mathbf{z},\boldsymbol{\theta},\mathbf{x},\mathbf{y})$ and $\mathbf{a}^i(\mathbf{W})$ collects the coefficients of $\boldsymbol{\omega}$ and $b^i$ is a constant. Let $\mathbf{e}_n$ denote the $n$-th unit vector in $\mathbb{R}^N$, and assume that the right-hand side is affine in the uncertainty, i.e., $b_m(\boldsymbol{\theta}) = b_m^0 + \boldsymbol{\beta}_m^\top\boldsymbol{\theta}$, while $\mathbf{a}_m$ and $\mathbf{e}_m$ do not depend on $\boldsymbol{\theta}$. Then:
\begin{itemize}
    \item for  constraints~\eqref{DRO_constraint1}, $\mathbf{a}^{n}(\mathbf{W}) = (\mathbf{w}_n,\, \mathbf{0},\, -\mathbf{e}_n,\, \mathbf{0})$ and $b^{n} = -0.5$, for $n = 1,\dots,N$;
    \item for  constraints~\eqref{DRO_constraint2}, $\mathbf{a}^{N+n}(\mathbf{W}) = (-\mathbf{w}_n,\, \mathbf{0},\, \mathbf{e}_n,\, \mathbf{0})$ and $b^{N+n} = -0.5$, for $n = 1,\dots,N$;
    \item for constraints~\eqref{DRO_constraint3}, $\mathbf{a}^{2N+m} = (\mathbf{0},\, -\boldsymbol{\beta}_m,\, \mathbf{a}_m,\, \mathbf{e}_m)$ and $b^{2N+m} = -b_m^0$, for $m = 1,\dots,M$;
\end{itemize}
Hence, the feasible set induced by these constraints is
\[
\Omega_C\triangleq
\left\{
\mathbf{W}\in\mathcal{W}:
\inf_{Q\in\mathcal{P}_{\boldsymbol{\omega}}(\rho)}
\mathbb{P}^Q\!\left[
\mathbf{a}^i(\mathbf{W})^{\top}\boldsymbol{\omega}+\mathbf{b}^i\le 0
\right]
\ge 1-\epsilon,\quad i=1,\dots,I
\right\}.
\]
\color{black}The lifted ambiguity set does not attempt to recover the unknown functional
dependence of the recourse decisions on the exogenous uncertainty. Rather, it
\emph{relaxes} the graph of this dependence by admitting distributions over
the full joint space of $(\mathbf{z},\boldsymbol{\theta},\mathbf{x},\mathbf{y})$,
without enforcing that
$(\mathbf{x},\mathbf{y})=H(\mathbf{z},\boldsymbol{\theta})$. Any deterministic
recourse mapping induces a graph-supported distribution that is a special case
of a joint distribution over the lifted space, so the construction is
conservative. This means that the adversary is granted more freedom than under the true
functional dependence. The following proposition makes this precise.
\begin{prop}[Conservatism of the Lifted Ambiguity Set]
\label{prop:conservatism}
Let $\mathcal{B}_\rho^H(\hat{P}_{\boldsymbol{\omega}})$ denote the subset of
$\mathcal{P}_{\boldsymbol{\omega}}(\rho)$ whose support is contained in the
graph $\mathcal{G}_H$, i.e.,
\[
\mathcal{B}_\rho^H(\hat{P}_{\boldsymbol{\omega}})
=\bigl\{Q\in\mathcal{P}_{\boldsymbol{\omega}}(\rho)
:\mathrm{supp}(Q)\subseteq\mathcal{G}_H\bigr\}.
\]
Then, for any measurable feasibility event $\mathcal{F}$,
\[
\inf_{Q\in\mathcal{P}_{\boldsymbol{\omega}}(\rho)}\mathbb{P}^Q(\mathcal{F})
\;\leq\;
\inf_{Q\in\mathcal{B}_\rho^H(\hat{P}_{\boldsymbol{\omega}})}\mathbb{P}^Q(\mathcal{F}).
\]
Consequently, if the chance constraint~\eqref{DRCC_compact} is satisfied over
$\mathcal{P}_{\boldsymbol{\omega}}(\rho)$, it is also satisfied over every
graph-supported distribution in $\mathcal{B}_\rho^H(\hat{P}_{\boldsymbol{\omega}})$.
\end{prop}
\begin{proof}
By definition, $\mathcal{B}_\rho^H(\hat{P}_{\boldsymbol{\omega}})\subseteq
\mathcal{P}_{\boldsymbol{\omega}}(\rho)$. Since the infimum of a function over a
set cannot exceed its infimum over a subset, the inequality follows.
\end{proof}

\begin{remark}
Note that $\mathbf{x}^s$ and $\mathbf{y}^s$ appear both as decision variables
in the training problem~\eqref{prescriptive_linear} and as components of the
lifted samples $\boldsymbol{\omega}^s$ that center the Wasserstein ball. At
the optimal solution, the policy-consistency
constraints~\eqref{prescrptive_linear_con1}--\eqref{prescrptive_linear_con2}
ensure that $\mathbf{x}^s$ is consistent with the thresholded output of
$\boldsymbol{\pi}_\mathbf{W}(\mathbf{z}^s)$, and the feasibility
constraint~\eqref{classical_feasibility_con3} guarantees that
$(\mathbf{x}^s,\mathbf{y}^s)$ is feasible for scenario $s$ by construction.
The lifted samples $\boldsymbol{\omega}^s$ therefore represent actual feasible
tuples consistent with the learned policy, and the Wasserstein ball provides a
robustness guarantee relative to the data distribution induced by training.
\end{remark}
\color{black}
Using the reformulation of individual Wasserstein distributionally robust
chance constraints presented in Proposition~1 of~\citep{chen2022}, and
introducing auxiliary variables $q_{is}\in\{0,1\}$, $r_{is}\ge 0$,
$t_i\in\mathbb{R}$, the feasible set $\Omega_C$ admits the following
representation:
\[
\Omega_C=
\left\{
\mathbf{W}\in\mathcal{W}:
\begin{array}{ll}
\epsilon S t_i-\displaystyle\sum_{s=1}^{S}r_{is}
\ge\rho S\left\|\mathbf{a}^i(\mathbf{W})\right\|_*,
& i=1,\ldots,I,\\[1.2em]
-\mathbf{a}^i(\mathbf{W})^\top{\boldsymbol{\omega}}^s
-\mathbf{b}^i+M_i q_{is}\ge t_i-r_{is},
& i=1,\ldots,I,\;s=1,\ldots,S,\\[0.8em]
M_i(1-q_{is})\ge t_i-r_{is},
& i=1,\ldots,I,\;s=1,\ldots,S,\\[0.8em]
q_{is}\in\{0,1\},\;r_{is}\ge 0,\;t_i\in\mathbb{R},
& i=1,\ldots,I,\;s=1,\ldots,S.
\end{array}
\right\}.
\]
Here, $\|\cdot\|_*$ denotes the norm dual to the ground norm used in the
Wasserstein distance, and $M_i$ is a sufficiently large positive constant.
This provides a tractable mixed-integer representation of the individual
Wasserstein distributionally robust chance constraints~\citep{chen2022}.

\subsection{Proposed Distributionally Robust Formulation}

We now specialize the compact reformulation of $\Omega_C$ to the three
families of individual chance constraints in \eqref{dro_chance}.
Introducing auxiliary variables
$q^{+}_{ns}\in\{0,1\}$, $r^{+}_{ns}\ge 0$,
$t^{+}_n\in\mathbb{R}$, $q^{-}_{ns}\in\{0,1\}$, $r^{-}_{ns}\ge 0$, and
$t^{-}_n\in\mathbb{R}$, the Wasserstein reformulation of chance constraints~\eqref{DRO_constraint1}-~\eqref{DRO_constraint2} is given by 
\begin{subequations}
\label{DRO_re1_plus}
\begin{align}
&\epsilon S t^{+}_n-\sum_{s=1}^{S} r^{+}_{ns}
\ge
\rho S\left\|[\mathbf{w}^n,-1]\right\|_*,
&& n=1,\ldots,N,
\label{eq:wzplus-1-new}
\\
&-(\mathbf{w}^n)^\top \mathbf{z}^s
+x_n^s+0.5
+M_n^{+}q^{+}_{ns}
\ge
t^{+}_n-r^{+}_{ns},
&& n=1,\ldots,N,\; s=1,\ldots,S,
\label{eq:wzplus-2-new}
\\
&M_n^{+}(1-q^{+}_{ns})
\ge
t^{+}_n-r^{+}_{ns},
&& n=1,\ldots,N,\; s=1,\ldots,S,
\label{eq:wzplus-3-new}
\\
&q^{+}_{ns}\in\{0,1\},\quad r^{+}_{ns}\ge 0,\quad t^{+}_n\in\mathbb{R},
&& n=1,\ldots,N,\; s=1,\ldots,S.
\label{eq:wzplus-4-new}\\
&\epsilon S t^{-}_n-\sum_{s=1}^{S} r^{-}_{ns}
\ge
\rho S\left\|[-\mathbf{w}^n,1]\right\|_*,
&& n=1,\ldots,N,
\label{eq:wzminus-1-new}
\\
&(\mathbf{w}^n)^\top \mathbf{z}^s
-x_n^s+0.5
+M_n^{-}q^{-}_{ns}
\ge
t^{-}_n-r^{-}_{ns},
&& n=1,\ldots,N,\; s=1,\ldots,S,
\label{eq:wzminus-2-new}
\\
&M_n^{-}(1-q^{-}_{ns})
\ge
t^{-}_n-r^{-}_{ns},
&& n=1,\ldots,N,\; s=1,\ldots,S,
\label{eq:wzminus-3-new}
\\
&q^{-}_{ns}\in\{0,1\},\quad r^{-}_{ns}\ge 0,\quad t^{-}_n\in\mathbb{R},
&& n=1,\ldots,N,\; s=1,\ldots,S.
\label{eq:wzminus-4-new}
\end{align}
\end{subequations}

Introducing auxiliary variables
$\hat q_{ms}\in\{0,1\}$, $\hat r_{ms}\ge 0$, and
$\hat t_m\in\mathbb{R}$, the Wasserstein reformulation of chance constraint \ref{DRO_constraint3} is
\begin{subequations}
\label{DRO_re2_new}
\begin{align}
&\epsilon S \hat t_m-\sum_{s=1}^{S} \hat r_{ms}
\ge
\rho S
\left\|[-\boldsymbol{\beta}^m,\mathbf{a}^m,\mathbf{e}^m]\right\|_*,
&& m=1,\ldots,M,
\label{eq:feas-1-new}
\\
&-(\mathbf{a}^m)^\top\mathbf{x}^s
-(\mathbf{e}^m)^\top\mathbf{y}^s
+(\boldsymbol{\beta}^m)^\top\boldsymbol{\theta}^s
+b_m^0+\hat M_m\hat q_{ms}
\ge
\hat t_m-\hat r_{ms},
&& m=1,\ldots,M,\; s=1,\ldots,S,
\label{eq:feas-2-new}
\\
&\hat M_m(1-\hat q_{ms})
\ge
\hat t_m-\hat r_{ms},
&& m=1,\ldots,M,\; s=1,\ldots,S,
\label{eq:feas-3-new}
\\
&\hat q_{ms}\in\{0,1\},\quad \hat r_{ms}\ge 0,\quad \hat t_m\in\mathbb{R},
&& m=1,\ldots,M,\; s=1,\ldots,S.
\label{eq:feas-4-new}
\end{align}
\end{subequations}
\color{black}
\begin{remark}[Training decisions as scenario-specific recourse]
\label{rem:xsy-decision-variables}
The variables $\mathbf{x}^s$ and $\mathbf{y}^s$ in
\eqref{DRO_re2_new} are scenario-specific decisions rather than
pre-computed observations. They remain coupled to the learned policy through
constraints~\eqref{prescrptive_linear_con1}--\eqref{prescrptive_linear_con2},
while constraint~\eqref{classical_feasibility_con3} ensures their feasibility
for each training scenario. Keeping $(\mathbf{x}^s,\mathbf{y}^s)$ endogenous
therefore allows the lifted training samples and the policy parameters
$\mathbf{W}$ to be optimized jointly. Pre-computing these decisions would
break this coupling and would no longer yield the same distributionally robust
feasibility construction.
\end{remark}

Combining the prescriptive objective \eqref{prescriptive_objectivefunction}
with the reformulations \eqref{DRO_re1_plus}--\eqref{DRO_re2_new} yields the
distributionally robust prescriptive learning model with feasibility guarantee.
The resulting formulation is a mixed-integer conic program. When the Wasserstein
ground norm is the $\ell_1$- or $\ell_\infty$-norm, the dual-norm constraints
are linear and the model reduces to a mixed-integer linear program. When the
ground norm is the $\ell_2$-norm, the dual-norm constraints are second-order cone
constraints, and the resulting model is a mixed-integer second-order cone problem.

\subsection{Illustrative Example under Distribution Shift} \label{example2}
We first provide a simple example to illustrate why distributional robustness
is useful when the test distribution differs from the training distribution.
Consider a binary decision problem with a single uncertain parameter
$\theta \in \mathbb{R}$ and one binary decision variable $x \in \{0,1\}$.
As in Subsection~\ref{example1}, we assume perfect information, i.e., the realized value of $\theta$ is used as the feature ($z=\theta$). This isolates the effect of the distribution shift between training and testing from that of forecast errors. In practice, $z$ would represent a forecast of $\theta$ available at decision time (see Remark~1).
Consider a linear policy of the form
$\pi(\theta)=w\theta+w_0.$
For a value of $\theta$, the underlying optimization problem is
\begin{subequations}
\begin{align}
\min_{x} \quad & -x \label{eq:toy_obj}\\
\text{s.t.}\quad & \theta+x\le 2, \label{eq:toy_feas}\\
& x\in\{0,1\}. \label{eq:toy_binary}
\end{align}
\end{subequations}
The objective encourages the decision-maker to choose $x=1$ whenever feasible.
Since constraint \eqref{eq:toy_feas} implies that $x=1$ is feasible only if
$\theta\le 1$, the optimal decision is
\[
x^*(\theta)=
\begin{cases}
1, & \theta\le 1,\\
0, & \theta>1.
\end{cases}
\]
Note that $x=0$ is always a feasible solution in this example. We consider
the following training samples:
\[
\{\theta^s_{\mathrm{train}}\}_{s=1}^{10}
=
\{0.00,\,0.10,\,0.20,\,0.30,\,0.45,\,0.65,\,0.80,\,0.90,\,0.95,\,1.00\},
\]
and set $z^s_{\mathrm{train}}=\theta^s_{\mathrm{train}}$. All training samples
satisfy $\theta^s_{\mathrm{train}}\le 1$, and therefore the optimal decision
is $x^{*,s}=1$ for every training instance. To mimic a shift in the
data-generating process, we evaluate the trained policies on the
upward-shifted test samples
\[
\{\theta^s_{\mathrm{test}}\}_{s=1}^{10}
=
\{0.10,\,0.20,\,0.35,\,0.50,\,0.70,\,0.90,\,1.02,\,1.08,\,1.15,\,1.25\},
\]
with $z^s_{\mathrm{test}}=\theta^s_{\mathrm{test}}$. The last four test
samples lie in the region $\theta>1$, where the only feasible decision is
$x=0$.

For comparison, we train three linear policies: classical predictive learning,
prescriptive learning, and distributionally robust prescriptive learning. In
the classical predictive approach, all supervised labels in the training set
are equal to one. Consequently, the learned policy is encouraged to predict
$1$ for all samples. Likewise, in the prescriptive formulation, the objective
favors $x^s=1$, and because $x^s=1$ is feasible for every training sample,
the model has no incentive to move the policy below the threshold near the
boundary $\theta=1$. Hence, both the classical and prescriptive approaches
learn policies that select $x=1$ throughout the training region and continue
to select $x=1$ for the shifted test samples, including those with $\theta>1$.

By contrast, the distributionally robust model accounts for local perturbations
around the empirical distribution of the training samples. Since several
training observations lie close to the feasibility boundary, namely
$\theta=0.90$, $0.95$, and $1.00$, the Wasserstein ambiguity set contains
nearby distributions that place probability mass beyond $\theta=1$. The
distributionally robust prescriptive formulation therefore favors a more
conservative linear policy that crosses the threshold before entering the
unsafe region. As a result, the distributionally robust policy may sacrifice
some objective value on training samples close to the boundary, but it
substantially improves out-of-sample feasibility under the shifted test
distribution.

\paragraph{Distributionally robust prescriptive reformulation.}
Because $z=\theta$ and there is no continuous recourse variable $y$, the
augmented random vector reduces to
\[
\omega=(\theta,x),
\]
where the $z$ and $\theta$ components coincide and the $y$ component is absent.
The three individual constraints to be robustified are
\begin{align}
&w\theta+w_0-x-0.5 \le 0, \label{eq:toy_upper_policy}\\
&-w\theta-w_0+x-0.5 \le 0, \label{eq:toy_lower_policy}\\
&\theta+x-2 \le 0. \label{eq:toy_feas_policy}
\end{align}
We use the $\ell_\infty$ ground norm in the Wasserstein metric, whose dual
norm is the $\ell_1$ norm. Hence,
\[
\|(w,-1)\|_1=|w|+1,\qquad
\|(-w,1)\|_1=|w|+1,\qquad
\|(1,1)\|_1=2.
\]
Let $\rho$ denote the Wasserstein radius and let $\epsilon$ denote the risk
level. Introducing binary variables $q_s^+,q_s^-,q_s^{\mathrm{f}}\in\{0,1\}$,
nonnegative variables $r_s^+,r_s^-,r_s^{\mathrm{f}}\ge 0$, and continuous
variables $t^+,t^-,t^{\mathrm{f}}\in\mathbb{R}$, the formulation becomes
\begin{subequations}
\begin{align}
&\epsilon S t^+ - \sum_{s=1}^S r_s^+
\ge \rho S (|w|+1), \label{eq:toy_dro_upper_main}\\
&-w\theta^s-w_0+x^s+0.5+M q_s^+
\ge t^+ - r_s^+,
\qquad s=1,\ldots,S, \label{eq:toy_dro_upper_margin}\\
&M(1-q_s^+)
\ge t^+ - r_s^+,
\qquad s=1,\ldots,S, \label{eq:toy_dro_upper_bigM}\\
&\epsilon S t^- - \sum_{s=1}^S r_s^-
\ge \rho S (|w|+1), \label{eq:toy_dro_lower_main}\\
&w\theta^s+w_0-x^s+0.5+M q_s^-
\ge t^- - r_s^-,
\qquad s=1,\ldots,S, \label{eq:toy_dro_lower_margin}\\
&M(1-q_s^-)
\ge t^- - r_s^-,
\qquad s=1,\ldots,S, \label{eq:toy_dro_lower_bigM}\\
&\epsilon S t^{\mathrm{f}} - \sum_{s=1}^S r_s^{\mathrm{f}}
\ge 2\rho S, \label{eq:toy_dro_feas_main}\\
&2-\theta^s-x^s+M q_s^{\mathrm{f}}
\ge t^{\mathrm{f}} - r_s^{\mathrm{f}},
\qquad s=1,\ldots,S, \label{eq:toy_dro_feas_margin}\\
&M(1-q_s^{\mathrm{f}})
\ge t^{\mathrm{f}} - r_s^{\mathrm{f}},
\qquad s=1,\ldots,S, \label{eq:toy_dro_feas_bigM}\\
&q_s^+,\,q_s^-,\,q_s^{\mathrm{f}}\in\{0,1\},\quad
r_s^+,\,r_s^-,\,r_s^{\mathrm{f}}\ge 0,\quad
t^+,\,t^-,\,t^{\mathrm{f}}\in\mathbb{R},
\qquad s=1,\ldots,S. \label{eq:toy_dro_domains}
\end{align}
\end{subequations}
The resulting distributionally robust training model minimizes the same
prescriptive objective as the prescriptive formulation, but replaces empirical
feasibility by the above distributionally robust constraints.

\begin{figure}[!htbp]
    \centering
    \includegraphics[width=0.72\textwidth]{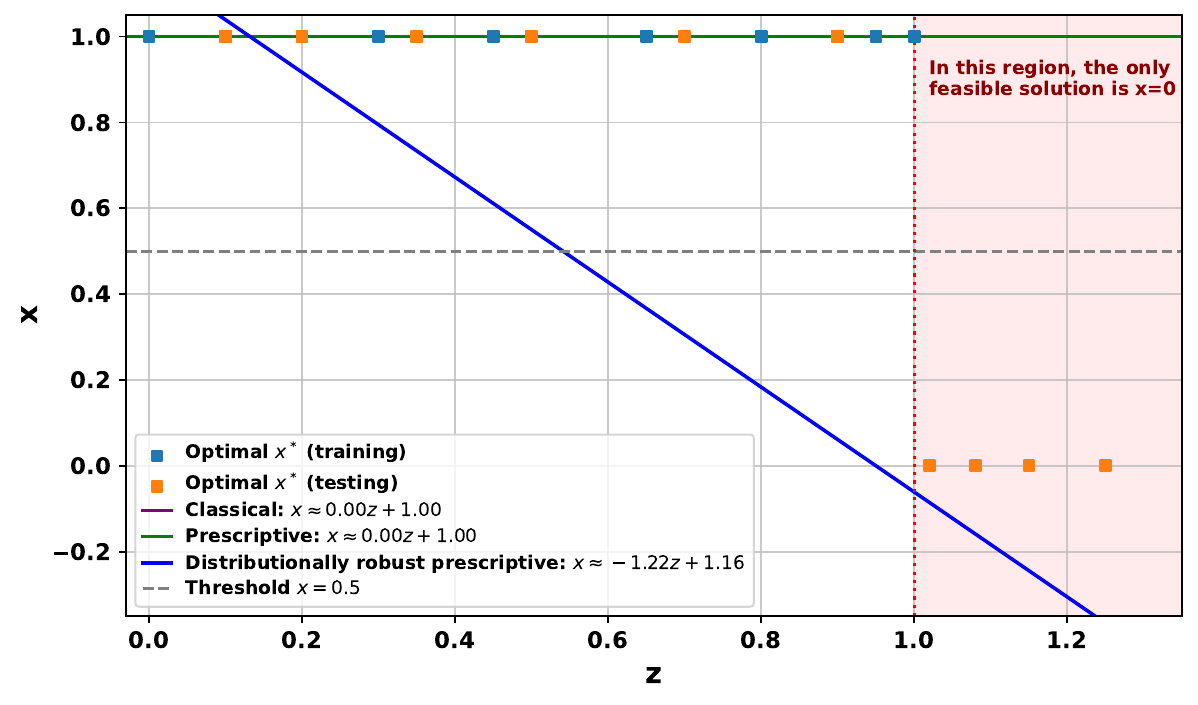}
    \caption{Learned linear policies under distribution shift. The classical
    predictive and prescriptive models both learn the constant policy
    $\pi(\theta)=1$ because all training samples satisfy $\theta\le 1$ and
    therefore have optimal decision $x^*=1$. Under the shifted test samples,
    these policies remain above the threshold and become infeasible when
    $\theta>1$. The distributionally robust policy shifts downward and
    prescribes $x=0$ for larger values of $\theta$, improving feasibility
    under distribution shift.}
    \label{fig:dro_toy_example}
\end{figure}

Figure~\ref{fig:dro_toy_example} compares the three learned linear policies.
The blue squares denote the optimal decisions for the training samples and the
orange squares denote the optimal decisions for the shifted testing samples.
In the training set, all realizations satisfy $\theta \le 1$, and hence the
optimal decision is always $x^*=1$. Therefore, both the classical predictive
and prescriptive models learn the constant policy $\pi(\theta)=1$, which
remains above the decision threshold $0.5$ over the entire domain. Note that
these two policies are identical in this example because, on the training data,
choosing $x=1$ is both optimal and feasible for every sample.

The shifted test set contains realizations with $\theta>1$. In this region,
highlighted in red in Figure~\ref{fig:dro_toy_example}, the only feasible
decision is $x=0$, since the constraint $\theta+x\le 2$ is violated whenever
$\theta>1$ and $x=1$. Nevertheless, the classical and prescriptive policies
continue to prescribe $x=1$ for these shifted test samples and therefore
become infeasible out of sample.

In contrast, the distributionally robust policy is no longer fitted only to
the empirical training samples. By enforcing feasibility over distributions in
a Wasserstein neighborhood of the empirical distribution, it anticipates
perturbations beyond the observed training support, especially near the
boundary $\theta=1$. As shown in Figure~\ref{fig:dro_toy_example}, the
distributionally robust policy tilts downward and crosses the threshold before
reaching the unsafe region. Consequently, it prescribes $x=0$ for shifted
test samples with larger values of $\theta$, thereby improving out-of-sample
feasibility at the expense of a more conservative policy on some training
realizations. This example illustrates the main role of the distributionally
robust formulation: it guards against feasibility failures caused by
distribution shifts that are not visible in the training data.

\section{Case Studies and Numerical Results}\label{results}
This section presents numerical analyses of the proposed prescriptive learning framework for binary decisions for two use cases, namely unit commitment and staff allocation problems. We consider three models for training as follows:
\begin{itemize}
    \item Model 1: Classical predictive learning (\ref{classical}) considering linear policy
    \item Model 2: Proposed prescriptive learning (\ref{prescriptive}) considering linear policy
    \item Model 3: Proposed distributionally robust prescriptive linear policy~\eqref{prescriptive_objectivefunction}, (\ref{DRO_re1_plus})-(\ref{DRO_re2_new})
\end{itemize}

In addition, we report the hindsight cost as a lower bound benchmark.
It is defined as the sample average of optimal costs obtained by solving the
corresponding deterministic problem for each test scenario individually.
The hindsight provides a meaningful
reference against which the quality of each learned policy can be assessed. The purpose of this experiment is to investigate whether incorporating the
unit commitment model directly into the learning problem improves the
operational quality and feasibility of the learned commitment policy, and
whether the distributionally robust chance-constrained formulation further
protects the policy against changes between the training and test
distributions. The code to reproduce the results is made available
in \cite{Pourahmadi}.

\subsection{Unit Commitment}
\label{sec:uc_case}
We first evaluate the proposed learning framework on a network-constrained
unit commitment problem. This application is particularly relevant because the
commitment decisions are binary, must be made before demand is fully observed,
and are strongly coupled through intertemporal and network constraints.
An inaccurate commitment policy may result not only in a higher
operating cost, but also in an infeasible dispatch caused by insufficient
generation capacity, limited ramping capability, or transmission congestion.

\subsubsection{Deterministic Formulation}
\label{sec:uc_deterministic}
Consider a power grid consisting of $N$ buses, $G$ generating units, and
$L$ transmission lines over a scheduling horizon of $T$ time periods.
The unit commitment problem includes both continuous and binary decision
variables. For each time period $t$, let
$\mathbf{p}_{t}=\{p_{1t},\ldots,p_{Gt}\}\in\mathbb{R}^{G}$
denote the power production of the generating units. Furthermore, let
$\mathbf{x}_{t}=\{x_{1t},\ldots,x_{Gt}\}\in\{0,1\}^{G}$ and
$\mathbf{u}_{t}=\{u_{1t},\ldots,u_{Gt}\}\in\{0,1\}^{G}$
denote the commitment and start-up decisions, respectively. Specifically,
$x_{gt}=1$ if generating unit $g$ is online at time $t$, and $u_{gt}=1$
if the unit starts up at time $t$.
For each generating unit $g$, the parameters $\underline{p}_{g}$ and
$\overline{p}_{g}$ denote its minimum and maximum active power generation
levels, respectively. The parameters $\overline{r}_{g}$ and
$\underline{r}_{g}$ represent its ramp-up and ramp-down limits, while
$\widetilde{r}_{g}$ denotes the ramping capability during start-up and
shut-down periods. The minimum up- and down-time requirements are denoted by
$\overline{\nu}_{g}$ and $\underline{\nu}_{g}$, respectively. The parameters
$c_{g}^{U}$, $c_{g}^{\mathrm{X}}$, and $c_{g}^{P}$ represent the start-up,
online, and variable production costs of generating unit $g$, respectively.
For each transmission line $\ell$, $h_{\ell n}$ denotes the power transfer
distribution factor with respect to bus $n$, and $\overline{e}_{\ell}$
denotes the transmission capacity. The demand at bus $n$ and time $t$ is
denoted by $d_{nt}$, and
$\mathbf{d}_{t}=\{d_{1t},\ldots,d_{Nt}\}\in\mathbb{R}^{N}$ collects the
nodal demands at time $t$.
Using this notation, the deterministic unit commitment problem is formulated
as follows:
\begin{subequations}
\label{eq:uc_deterministic}
\begin{align}
\min_{\mathbf{p},\mathbf{x},\mathbf{u}}
\quad &
\sum_{t=1}^{T}\sum_{g=1}^{G}
\left(
c_{g}^{U}u_{gt}+c_{g}^{\mathrm{X}}x_{gt}+c_{g}^{P}p_{gt}
\right)
\label{eq:uc_det_objective}
\\
\text{s.t.}\quad
&
-x_{g(t-1)}+x_{gt}-x_{g\tau}\leq 0,
&& \forall g,t,\ \tau\in\{t,\ldots,t+\overline{\nu}_{g}-1\},
\label{eq:uc_det_minup}
\\
&
x_{g(t-1)}-x_{gt}+x_{g\tau}\leq 1,
&& \forall g,t,\ \tau\in\{t,\ldots,t+\underline{\nu}_{g}-1\},
\label{eq:uc_det_mindown}
\\
&
-x_{g(t-1)}+x_{gt}-u_{gt}\leq 0,
&& \forall g,t,
\label{eq:uc_det_startup}
\\
&
\underline{p}_{g}x_{gt}
\leq p_{gt}
\leq \overline{p}_{g}x_{gt},
&& \forall g,t,
\label{eq:uc_det_capacity}
\\
&
p_{gt}-p_{g(t-1)}
\leq
\overline{r}_{g}x_{g(t-1)}
+\widetilde{r}_{g}\left(1-x_{g(t-1)}\right),
&& \forall g,t,
\label{eq:uc_det_rampup}
\\
&
p_{g(t-1)}-p_{gt}
\leq
\underline{r}_{g}x_{gt}
+\widetilde{r}_{g}\left(1-x_{gt}\right),
&& \forall g,t,
\label{eq:uc_det_rampdown}
\\
&
\sum_{g=1}^G p_{gt}
\geq
\sum_{n=1}^N d_{nt},
&& \forall t,
\label{eq:uc_det_balance}
\\
&
-\overline{e}_{\ell}
\leq
\sum_{n=1}^{N}
h_{\ell n}
\Big(
\sum_{g\in\mathcal{G}_n} p_{gt}-d_{nt}
\Big)
\leq
\overline{e}_{\ell},
&& \forall \ell,t,
\label{eq:uc_det_transmission}
\\
&
x_{gt},u_{gt}\in\{0,1\},
&& \forall g,t.
\label{eq:uc_det_domain}
\end{align}
\end{subequations}
Objective~\eqref{eq:uc_det_objective} minimizes the total operating cost,
consisting of the start-up cost $c_{g}^{U}u_{gt}$, the online cost
$c_{g}^{X}x_{gt}$, and the variable production cost
$c_{g}^{P}p_{gt}$. Constraints~\eqref{eq:uc_det_minup} and
\eqref{eq:uc_det_mindown} impose the minimum up- and down-time requirements.
In particular, once a generating unit is switched on, it must remain online
for at least $\overline{\nu}_{g}$ consecutive periods. Similarly, once it is
switched off, it must remain offline for at least
$\underline{\nu}_{g}$ consecutive periods.
Constraint~\eqref{eq:uc_det_startup} links the commitment and start-up
variables. It ensures that $u_{gt}=1$ whenever generating unit $g$
transitions from the offline state at time $t-1$ to the online state at time
$t$. Constraint~\eqref{eq:uc_det_capacity} enforces the minimum and maximum
generation limits conditional on the commitment status of the unit.
Constraints~\eqref{eq:uc_det_rampup} and
\eqref{eq:uc_det_rampdown} impose the ramp-up and ramp-down restrictions
between consecutive periods. These constraints distinguish between normal
ramping operation and the larger ramping capability allowed during start-up
or shut-down periods. Constraint~\eqref{eq:uc_det_balance} requires the
total generation to cover the total electricity demand in each time period.
Finally, constraint~\eqref{eq:uc_det_transmission} limits the power flow on
each transmission line using the power transfer distribution factors
$h_{\ell n}$ and the corresponding line capacity $\overline{e}_{\ell}$. For each bus $n$, $\mathcal{G}_n \subseteq \{1,\ldots,G\}$ denotes the set of generating units located at bus $n$.
Constraint~\eqref{eq:uc_det_domain} defines the commitment and start-up
variables as binary.

%

\subsubsection{Experimental setup}
\label{sec:uc_setup}
The numerical experiment uses a three-bus system with three generating
units, three loads, and three transmission lines, whose parameters are
adapted from standard unit commitment benchmarks
\cite{Carrion2006, Morales2013}. The scheduling
horizon contains $T=24$ hourly periods. Each generator is located at a
different bus, and each bus contains one load.
The maximum generation capacities are
$(0.55,\;0.40,\;0.20)$ p.u., while the minimum generation levels are
$(0.05,\;0.05,\;0.00)$ p.u.
The hourly ramp limits are $(0.30,\;0.25,\;0.40)$ p.u./h, and the minimum up-times
and down-times are $(2,\;2,\;1)$ periods. The variable production costs are
$(15,\;40,\;8)$ $\$/(\text{p.u.}\cdot \text{h})$, the online costs are $(1,\;5,\;1)$ \$/h, and the
start-up costs are $(5,\;20,\;3)$ \$ for generators 1--3, respectively.
The three loads are distributed across the buses using fixed shares
$(0.50,\;0.30,\;0.20)$. Consequently, the uncertainty is represented by a
single normalized system-demand trajectory, while its spatial
distribution remains fixed. The three transmission lines connect buses
1--2, 1--3, and 2--3, each with a capacity of $0.60$ p.u.
The combined maximum capacity of generators 1 and 3 is
$\overline{p}_1+\overline{p}_3=0.75$ p.u. Generator 2 is therefore required whenever demand exceeds
$0.75$ p.u. It is also the most expensive generator, with variable,
online, and start-up costs of $40$ $\$/(\text{p.u.}\cdot \text{h})$, $5$ \$/h, and $20$ \$,
respectively.
This capacity threshold plays an important role in the
distribution shift experiment.

Out-of-sample performance is evaluated by fixing the learned commitment
schedule and solving the resulting economic dispatch problem, augmented
with a load-shedding slack variable penalized at $c^{\mathrm{ls}} =
1{,}000$ $\$/(\text{p.u.}\cdot \text{h})$, a standard value used in the power systems literature
to represent the value of lost load \citep{Conejo2010, Morales2013}.
Note that load shedding is not part of the training problems.
Total cost is then decomposed into production cost, comprising variable
production, online, and start-up costs, and load-shedding cost.
All costs and unmet energies are totals over the $30$ trajectories of each test set.
We generate $30$ training trajectories, $30$ validation trajectories, and two sets of $30$ test trajectories. Each trajectory of realized demand is generated as
\[
d_t = \min\big\{\max\{\kappa\,\bar d_t + \eta_t,\;0.10\},\;0.99\big\}, \qquad t = 1,\ldots,24,
\]
where $\bar d_t$ is a common 24-hour base-demand profile with a peak of $0.90$~p.u.\ in the evening, $\kappa$ is a scaling factor, and $\eta_t$ is independent uniform noise. Each dataset consists of three demand levels with $10$ trajectories each: a high level with $\eta_t\sim\mathcal{U}[-0.03,0.03]$, a medium level with $\eta_t\sim\mathcal{U}[-0.05,0.05]$, and a low level with $\eta_t\sim\mathcal{U}[-0.03,0.03]$. The scaling factors are given in Table~\ref{tab:uc_data}.

The training, validation, and close test sets are drawn independently from the same distribution, with $\kappa\in\{0.83,\,0.62,\,0.45\}$ for the high, medium, and low levels. The training set is used to train the models, the validation set is used only to tune the hyperparameters $(\rho,\epsilon)$ of the distributionally robust model, and the close test set is used only for the final out-of-sample evaluation. With these scaling factors, generator~2 is needed only on a few peak hours of the high-demand training days, where the realized demand slightly exceeds the $0.75$~p.u.\ threshold. The far test set is drawn from a different, higher-demand distribution with $\kappa\in\{0.93,\,0.85,\,0.78\}$, so that the peak demand straddles the generator~2 commitment threshold.

\begin{table}[!htbp]
\centering
\caption{Scaling factors $\kappa$ of the demand distributions ($10$ trajectories per level).}
\label{tab:uc_data}
\begin{tabular}{lccc}
\toprule
Dataset & High & Medium & Low \\
\midrule
Training, validation, close test & $0.83$ & $0.62$ & $0.45$ \\
Far test & $0.93$ & $0.85$ & $0.78$ \\
\midrule
Noise $\eta_t$ & $\mathcal{U}[-0.03,0.03]$ & $\mathcal{U}[-0.05,0.05]$ & $\mathcal{U}[-0.03,0.03]$ \\
\bottomrule
\end{tabular}
\end{table}

The realized demand is not known when the commitment decisions are made. The features are therefore day-ahead demand forecasts, obtained by adding an independent Gaussian forecast error with standard deviation $0.02$~p.u.\ to the realized demand in each hour (mean absolute forecast error $0.015$~p.u.). The policy observes only the forecasts, whereas the realized demand enters the training problems through the power-balance and line constraints (and the oracle labels of the classical model) and is used out of sample to dispatch and evaluate the schedules.

\begin{figure}[!htbp]
  \centering
  \includegraphics[width=0.52\textwidth]{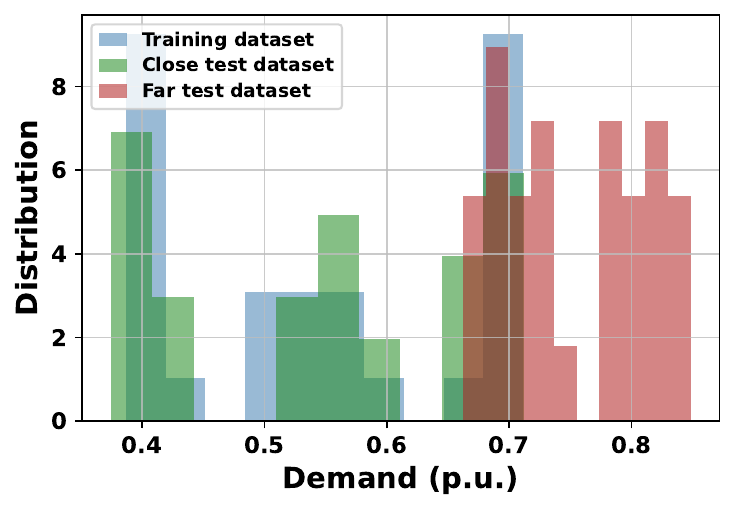}
  \caption{Demand forecast distributions for hour~18 across the three datasets.
    Wasserstein distances from the training set:
    $W_1(\text{training}, \text{close test}) = 0.017$ and
    $W_1(\text{training}, \text{far test}) = 0.185$.}
  \label{fig:demand_hour18}
\end{figure}

For each hour $t\in\{1,\ldots,24\}$, this yields one training distribution and two out-of-sample test distributions. We quantify the distance between the training set and each test set using the Wasserstein-1 metric computed on the forecasts. Since the close test set is drawn from the same distribution as the training set, the Wasserstein distance is small, whereas it is substantially larger for the far test set.
Figure~\ref{fig:demand_hour18} illustrates the per-hour forecast
distributions for hour~18, a representative peak hour in which the
distributional shift is clearly visible: training and close-test demand
concentrates mostly below the threshold (dashed line), with only a few
training realizations above it, while far-test demand
spreads across and above it. 

\subsubsection{Results}
\label{sec:uc_results}
We compare classical predictive learning, prescriptive learning, and distributionally robust prescriptive learning, and consider one linear policy per generator and hour. The formulations of both the prescriptive and the distributionally robust prescriptive unit commitment models are provided in Appendix~\ref{app:uc_models}. Figure~\ref{fig:uc_results} reports the out-of-sample performance of the
three learning models on the close and far test datasets, decomposed into
production cost in the top row and unmet demand in the middle row, with total
cost in the bottom row. The green dashed line is the hindsight benchmark,
which provides a lower bound on achievable total cost.

\paragraph{Close test.}
On the close test set, drawn from the same distribution as the training data, the
hindsight benchmark is approximately 4{,}900\$. The classical model incurs a total cost
of 5{,}093\$, of which 284\$ arises from load shedding (0.28 p.u.$\cdot$h of unmet
demand). Its oracle labels commit generator~2 only on the few training hours in which
it is strictly needed, so the regression learns to (almost) never commit it, and the
policy under-commits on high-demand hours. The prescriptive model nearly eliminates load
shedding (0.01 p.u.$\cdot$h, 10\$), because it must cover the realized demand in every
training trajectory and therefore learns to commit generator~2 when the forecast is
high. This comes at a higher production cost of 5{,}673\$, giving a total cost of
5{,}683\$. The distributionally robust prescriptive model has no load shedding and the
highest total cost of 6{,}709\$. This premium reflects the conservative nature of the
robust policy: by committing generator~2 more aggressively to hedge against the far
distribution, it over-commits in scenarios where demand remains low.

\paragraph{Far test.}
The advantage of the prescriptive and distributionally robust models becomes evident on
the far test set, whose demand distribution straddles the generator~2 commitment
threshold of $0.75$~p.u. The classical model incurs a total cost of 11{,}023\$, of which
4{,}850\$ is load-shedding cost (4.85 p.u.$\cdot$h of unmet demand): having learned a
policy that (almost) never commits generator~2, it cannot cover demand in the far-test
scenarios where generator~2 is indispensable. The prescriptive model reduces the unmet
energy to 1.99 p.u.$\cdot$h and the total cost to 9{,}201\$ ($16.5\%$ below the
classical model), since its policy commits generator~2 when the forecast is high.
However, this policy is calibrated to the training regime, in which generator~2 is
needed only on a few peak hours, and does not anticipate the higher demand levels of the
far test. The distributionally robust prescriptive model achieves the lowest total cost
of 8{,}726\$, $5.2\%$ below the prescriptive model and $20.8\%$ below the classical
model. Its unmet demand of 0.59 p.u.$\cdot$h is $70\%$ lower than that of the
prescriptive model and $88\%$ lower than that of the classical model, at the expense of
a moderately higher production cost of 8{,}135\$ ($13\%$) compared to 7{,}208\$.
This trade-off is precisely the intended effect of the
distributional robustification. By enlarging the effective training
distribution to cover demand scenarios near and above the commitment
threshold, the distributionally robust prescriptive model learns to commit
generator~2 when warranted, thereby avoiding most infeasible dispatches.

\begin{figure}[!htbp]
  \centering
  \includegraphics[width=\textwidth]{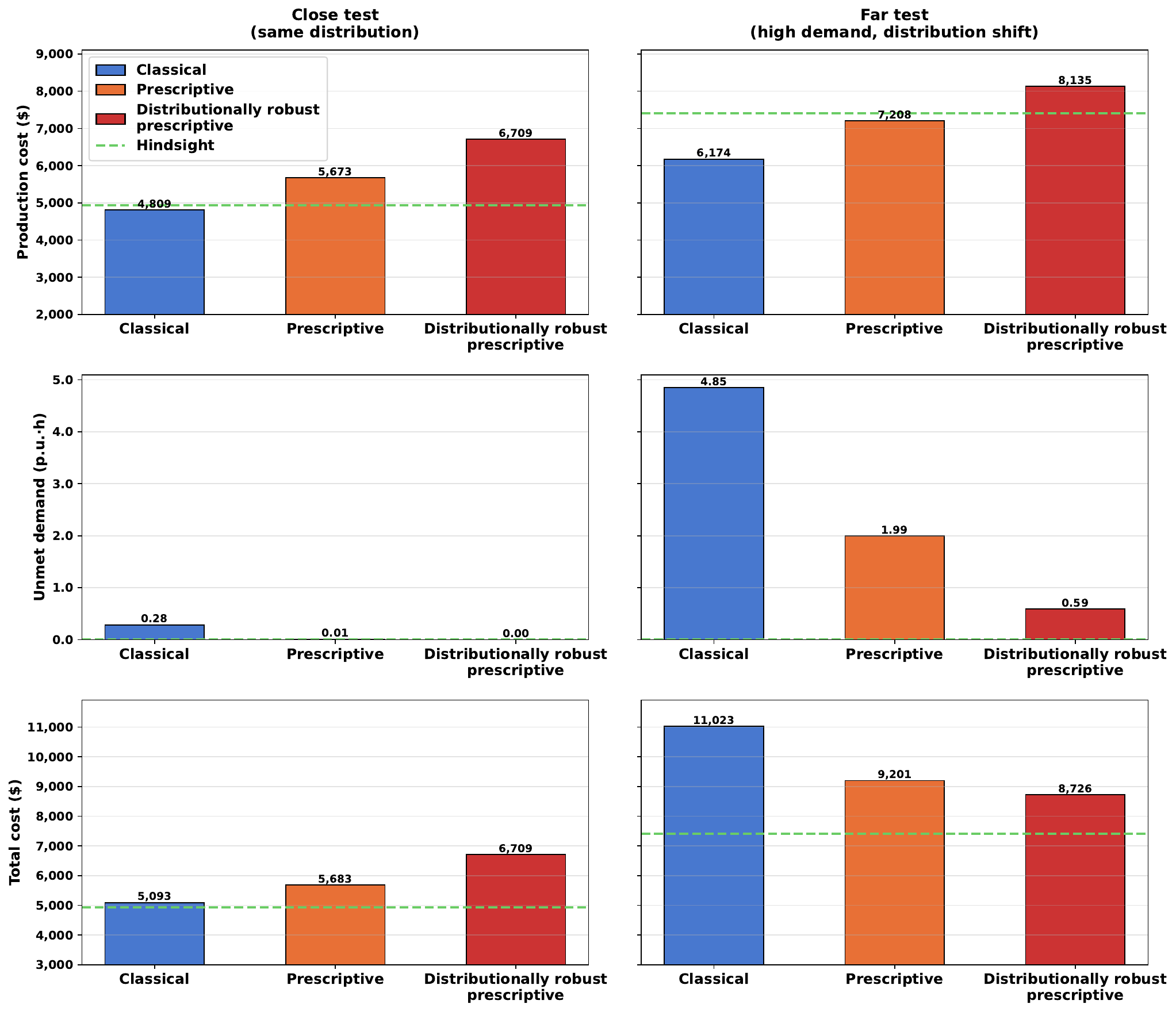}
  \caption{Out-of-sample results, totals over the $30$ trajectories of each test set:
    production cost
    (comprising variable production, online, and start-up costs; top row),
    unmet demand (middle row), and total cost (bottom row).
    Left column: close test (same distribution as training).
    Right column: far test (distribution shift from training).
    The green dashed line is the hindsight benchmark.}
  \label{fig:uc_results}
\end{figure}

\subsection{Staff Allocation}

\label{sec:sa_case}
We next evaluate the proposed framework on a shift-based staff allocation
problem. In this setting, a decision maker must determine a binary work
schedule indicating whether each employee is assigned to a given shift, so
that minimum staffing levels are met while minimizing total labor costs.
This problem is representative of a large class of pure-binary optimization
problems (i.e., those with no continuous recourse) in which the key
uncertainty is the shift demand. 

\subsubsection{Deterministic Formulation}
\label{sec:sa_deterministic}
Let $\mathcal{E}=\{1,\ldots,E\}$ denote the set of employees and
$\mathcal{J}=\{1,\ldots,J\}$ the set of work shifts. For each employee
$e\in\mathcal{E}$, the parameter $c_{e}$ denotes the unit cost per work shift
assigned, and $\underline{s}_{e}$ and $\overline{s}_{e}$ denote the
minimum and maximum number of work shifts that employee $e$ may be scheduled
for, respectively. Each work shift $j\in\mathcal{J}$ requires at least $d_{j}$
staff members. The binary decision variable $x_{ej}\in\{0,1\}$ equals one
if employee $e$ is assigned to work shift $j$ and zero otherwise. The
deterministic staff allocation problem is formulated as
\begin{subequations}
\label{eq:sa_deterministic}
\begin{align}
\min_{\mathbf{x}}
\quad &
\sum_{e=1}^{E}\sum_{j=1}^{J}
c_{e}\,x_{ej}
\label{eq:sa_det_objective}
\\
\text{s.t.}\quad
&
\sum_{e=1}^{E} x_{ej}
\geq d_{j},
&& \forall j,
\label{eq:sa_det_demand}
\\
&
\underline{s}_{e}
\leq
\sum_{j=1}^{J} x_{ej}
\leq \overline{s}_{e},
&& \forall e,
\label{eq:sa_det_shiftlimits}
\\
&
x_{ej}\in\{0,1\},
&& \forall e,j.
\label{eq:sa_det_domain}
\end{align}
\end{subequations}
Objective~\eqref{eq:sa_det_objective} minimizes the total staffing cost.
Constraint~\eqref{eq:sa_det_demand} ensures that each work shift $j$ is
covered by at least $d_{j}$ employees.
Constraint~\eqref{eq:sa_det_shiftlimits} enforces the individual workload
limits: each employee must be scheduled for no fewer than
$\underline{s}_{e}$ and no more than $\overline{s}_{e}$ work shifts.
Constraint~\eqref{eq:sa_det_domain} imposes the binary structure of the
assignment decisions.



\subsubsection{Experimental Setup}
\label{sec:sa_setup}

The setting is representative of shift-based staffing in hospitals, call
centers and other service systems, where the schedule is fixed in advance on
the basis of demand forecasts
\citep{ernst2004staff,vandenbergh2013personnel,green2007coping,defraeye2016staffing,kim2015two}.
The numerical experiment considers a workforce of $E=40$ employees to be
scheduled across $J=20$ shifts.
The unit staffing cost per employee-shift assignment $c_{e}$ is drawn as an
integer between $8$ and $15$ (in \$), and the maximum number of shifts each
employee may be assigned to, $\bar{s}_{e}$, is drawn from $\{2,\dots,5\}$,
yielding a total workforce capacity of $\sum_{e}\bar{s}_{e}=143$ assignment
slots.
All employees have a minimum workload of $\underline{s}_{e}=0$.

The demand for shift $j$ in scenario $i$ is modeled as
$d_{ij} = u_{ij}\,\bar{d}_{j}$,
where $\bar{d}_{j}\in\{6,\dots,12\}$ is the maximum staffing requirement
of shift $j$ (with $\sum_j\bar d_j=171$) and $u_{ij}\in(0,1]$ is the realized
demand utilization factor.
The realized demand is not known when the schedule is fixed; instead, the
decision maker observes a forecast $f_{ij}$ of the utilization, and
$u_{ij} = f_{ij} + \sigma\,\eta_{ij},\eta_{ij}\sim\mathcal{N}(0,1)$
drawn independently for each shift $j$ and scenario $i$, with forecast-error
standard deviation $\sigma=0.05$ ($u_{ij}$ is clipped to $[0.01,1]$).
The contextual feature vector $\mathbf{z}_{i}=(f_{i1},\dots,f_{iJ})\in\mathbb{R}^{J}$
collects the per-shift demand forecasts and serves as the sole input to the
linear assignment policy; the realized demand $d_{ij}$ enters training only
through the coverage constraints (and the oracle labels of the classical
model) and is used out of sample only to evaluate the schedules.
Out-of-sample performance is evaluated by fixing the learned binary schedule
$x_{ej}$, obtained from the forecasts, and computing the total cost as the sum
of staffing cost and demand shortage cost against the realized demand; any
unmet demand in shift $j$ is penalized at
$c^{\mathrm{ls}} = 500$ \$ per unit-shift shortage.
Costs are reported as means per test scenario.

We consider $S\in\{50,\,100,\,200,\,500\}$ training scenarios; the validation
set contains $200$ scenarios and each of the two test sets contains $1{,}000$
scenarios.
The training and close-test scenarios are drawn from a low-demand setting,
with $f_{ij}\!\sim\!\mathcal{U}(0.25,\,0.55)$ sampled independently for each
shift $j$, corresponding to a mean shift utilization of $40\%$ of peak
demand.
The far-test scenarios are drawn from a high-demand setting,
$f_{ij}\!\!\sim\!\mathcal{U}(0.65,\,0.90)$, raising the mean utilization to
$77.5\%$ and pushing aggregate demand requirements toward the total workforce
capacity.
We quantify the distributional shift via the Wasserstein-1 distance computed
between the training set and each test set per shift.
The close test yields near-zero Wasserstein distances (mean $0.005$), while the far test yields
substantially larger distances (mean $0.376$), reflecting the systematic upward
shift in demand levels.

The hyperparameters of the distributionally robust model, i.e., the Wasserstein
radius $\rho$ and the risk level $\varepsilon$, are selected by grid search
over the validation scenarios, explained in
Section~\ref{sec:uc_setup}.
The radius is parameterized as $\rho = \alpha\,\bar{W}_{1}$, where
$\bar{W}_{1}$
is the mean of the per-shift Wasserstein-1 distances between the
training empirical distribution  and the far-test
empirical distribution, each computed
over the scalar demand forecast $f_{ij}$, and
$\alpha\in\{0.05,\,0.03,\,0.02,\,0.01,\,0.005\}$.
The risk level is searched over $\varepsilon\in\{0.10,\,0.20,\,0.30\}$.
The hyperparameter combination that minimizes total validation cost is
retained for out-of-sample evaluation; for every sample size this yields
$\alpha=0.03$ ($\rho=0.011$) and $\varepsilon=0.30$.


\subsubsection{Results}
\label{sec:sa_results}

\begin{figure}[!htbp]
  \centering
  \includegraphics[width=\textwidth]{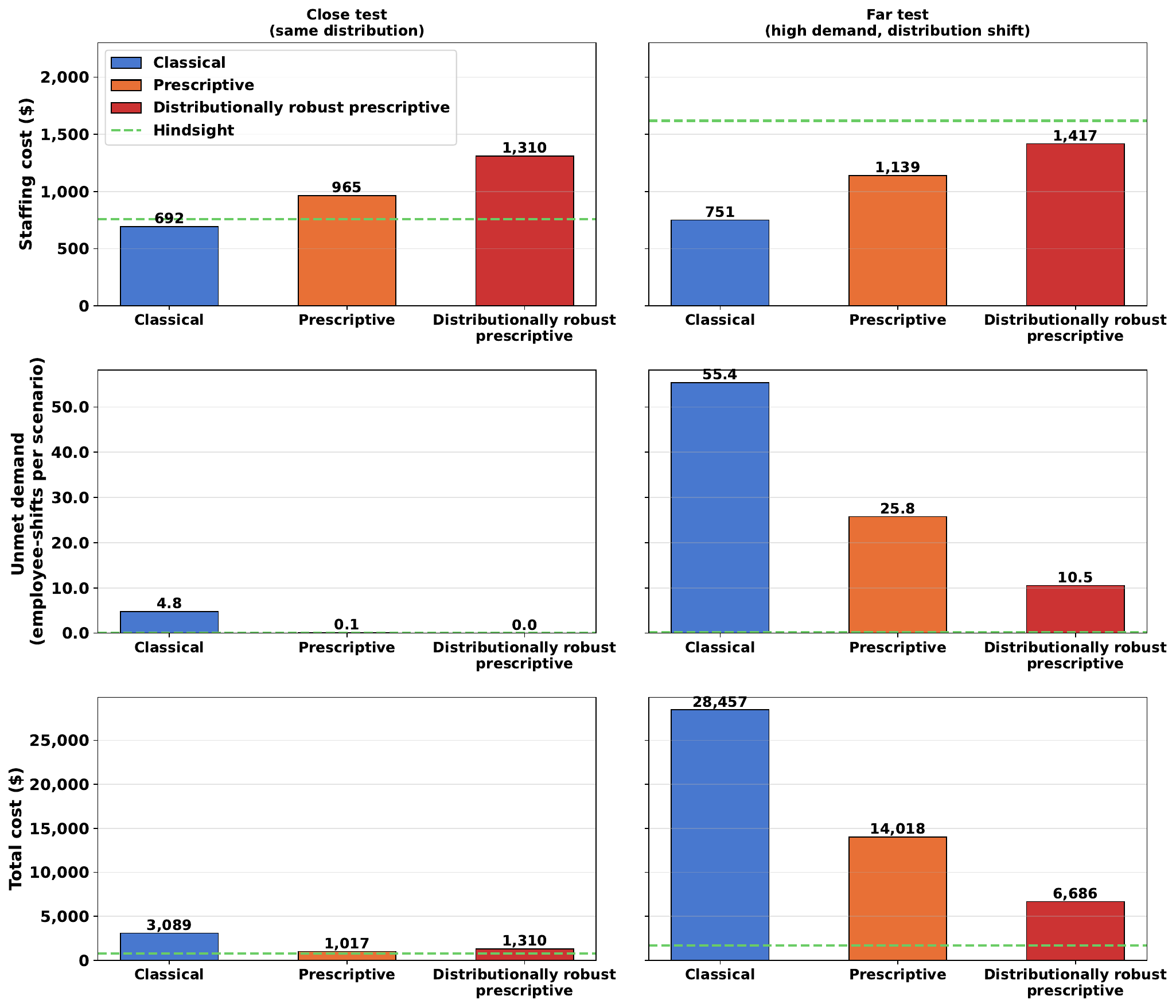}
  \caption{Out-of-sample cost comparison for the staff allocation use case
with $S=100$ training scenarios (means per test scenario).
Costs are decomposed into staffing cost (top), unmet demand in employee-shifts per scenario (middle),
and total cost (bottom) for the close test set, drawn from the same
distribution as the training data, and the far test set, which reflects a
high-demand distribution shift.
The green dashed line indicates the hindsight benchmark.}
\label{fig:sa_results}
\end{figure}

Figure~\ref{fig:sa_results} reports the out-of-sample costs of the three
learning models for $S=100$ training scenarios, decomposed into staffing cost
(top row) and unmet demand (middle row), with total cost in the bottom
row.
The green dashed line is the hindsight benchmark, which provides a lower bound on the achievable total cost.

\paragraph{Close test.}
On the close test set, drawn from the same distribution as the training data,
the classical model incurs a total cost of 3,089\$, of which 2,396\$ arises from
demand shortage (4.8 unmet employee-shifts per scenario).
This is the limitation of predictive learning discussed in
Section \ref{sec:Method}. The policy is trained to reproduce optimal schedules,
which staff each shift exactly to its requirement and therefore lie on the
boundary of the coverage constraints. Small prediction errors of the linear
policy then leave shifts understaffed. In addition, the large number of
employees with similar costs gives many alternative optimal schedules, so the
labels are inconsistent across scenarios; as a result, the classical policy
assigns fewer employees than the oracle schedules themselves, with a staffing
cost below the hindsight benchmark.
The prescriptive model achieves the lowest total cost of 1,017\$, with almost no
shortage (52\$, or 0.1 unmet employee-shifts). Since it enforces the coverage
constraint in every training scenario rather than imitating labels, it learns a feasible policy within the linear
policy class.
The distributionally robust prescriptive model has no shortage, at the expense of
a higher staffing cost of 1,310\$. As in the illustrative example of
Section \ref{section3}, the Wasserstein ambiguity set leads to a more
conservative policy, which assigns more employees than necessary when demand
stays low.

\paragraph{Far test.}
The advantage of distributional robustness becomes pronounced on the far test,
where demand levels systematically exceed those seen during training.
The classical model, whose policy was fitted on low-demand oracle solutions,
prescribes schedules that are grossly inadequate under the higher
demands.
It incurs a shortage cost of 27,706\$ and a total cost of 28,457\$,
representing a 1,589\% increase relative to the hindsight benchmark of 1,685\$.
The prescriptive model reduces the shortage cost to 12,879\$ and total cost
to 14,018\$.
The joint optimization of policy parameters and assignment decisions yields a
better staffing policy than the classical model, but the
in-sample demand coverage constraints cannot extrapolate to the unobserved
high-demand setting, and large shortages persist.
The distributionally robust prescriptive model achieves a staffing cost of
1,417\$, a shortage cost of 5,269\$, and a total cost of 6,686\$.
This represents a reduction in total cost of $76.5\%$ relative to the
classical model and $52.3\%$ relative to the prescriptive model,
while the shortage cost is reduced by $81\%$ compared to the classical
approach.
By expanding the effective training distribution through the Wasserstein
ambiguity set, the robust model learns to assign more employees across all
shifts, thereby absorbing the increased demands with substantially
fewer shortfalls.

\begin{figure}[!htbp]
  \centering
  \includegraphics[width=\textwidth]{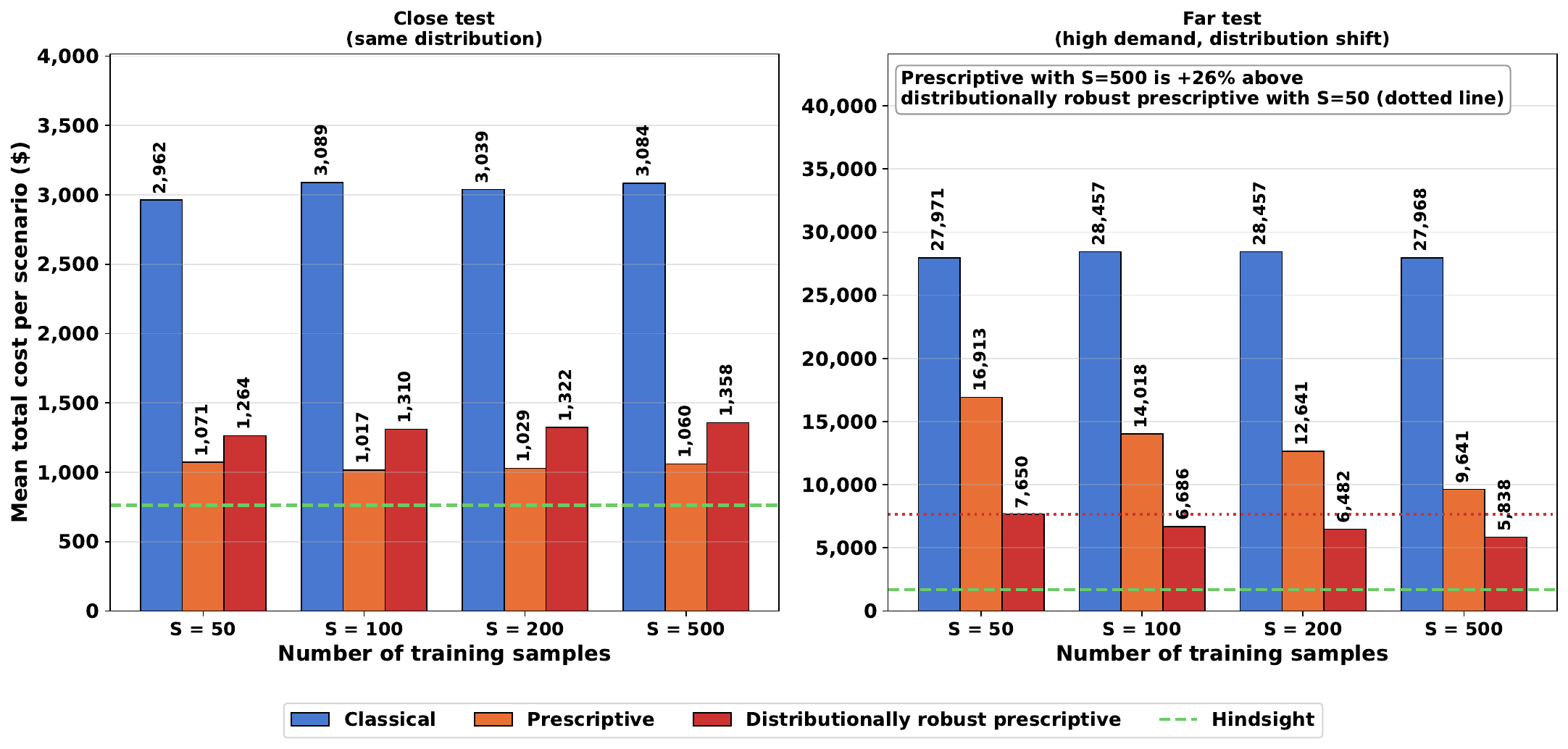}
  \caption{Mean out-of-sample total cost of the staff allocation use case as a
function of the number of training scenarios $S$, for the close test (left) and
the far test (right).
The green dashed line indicates the hindsight benchmark; the dotted line marks
the far-test cost of the distributionally robust model trained on $S=50$
scenarios.}
\label{fig:sa_sample_size}
\end{figure}

\paragraph{Sample size.}
Figure~\ref{fig:sa_sample_size} repeats the experiment for
$S\in\{50,\,100,\,200,\,500\}$; the ranking of the three models is the same for
every sample size.
The classical model does not improve with more data (about 3,000\$ on the close
test and 28,000\$ on the far test), since its error stems from imitating
hindsight-optimal labels rather than from estimation noise.
The prescriptive model becomes more conservative as $S$ grows: its far-test
cost falls from 16,913\$ ($S=50$) to 9,641\$ ($S=500$), while its close-test
cost remains between 1,017\$ and 1,071\$.
The distributionally robust model also benefits from more data under the shift,
with its far-test cost falling from 7,650\$ to 5,838\$, while its close-test
cost rises by only $7\%$ (from 1,264\$ to 1,358\$).
Importantly, additional samples do not close the gap between the two
prescriptive models. Even with $S=500$, the prescriptive model is $26\%$ more
expensive on the far test than the distributionally robust model trained on only
$S=50$ scenarios, and $65\%$ more expensive than the robust model trained on the
same $500$ scenarios.
More training data does not reveal the shift, whereas
the Wasserstein ambiguity set anticipates it explicitly.
The close-test cost of the prescriptive model does not decrease with $S$. Since the policy
observes only the forecast, its safety buffer is determined by the largest
forecast error among the $S$ training scenarios, which grows with $S$ and
eventually exceeds the cost-optimal service level.

\section{Conclusion}\label{sec:Conclusion}
This paper addressed the problem of binary decision-making under uncertainty, where the goal is to learn a policy that maps contextual features to binary decisions without solving an optimization problem at the time of decision making. Existing predictive learning approaches train policies to reproduce optimal decisions from labeled data, but lack knowledge of counterfactual outcomes and provide no guarantees of feasibility. This limitation becomes especially severe under distributional shift. We proposed a prescriptive learning framework that overcomes these limitations by integrating linear decision rules directly into the training optimization problem, enabling the policy to evaluate the full space of feasible alternatives and their degrees of suboptimality during training. We further extended the framework to a distributionally robust chance-constrained formulation using a Wasserstein ambiguity set, which provides feasibility guarantees under distributional uncertainty and introduces a tunable risk level to balance robustness against solution quality.
Numerical experiments on a unit commitment and a staff allocation problem demonstrate that the proposed framework consistently outperforms classical predictive learning under distributional shift. The distributionally robust model incurs a moderate increase in operational cost when evaluated on the training distribution, but reduces total cost by up to 76.5\% relative to the classical approach under distribution shift, driven by a dramatic reduction in infeasibility penalties. These results reinforce a central finding that aligning the learning objective with feasibility robustness is at least as important as minimizing predictive error.

Several directions remain open for future work. The linear policy parameterization, while interpretable and computationally convenient, may not capture complex nonlinear relationships between features and optimal decisions; extending the framework to nonlinear or tree-based policies is a natural next step. The training formulation currently scales linearly in the number of scenarios in terms of binary variables, which constrains the feasible sample size for large-scale problems; developing decomposition or approximation schemes to handle larger datasets would broaden the framework's applicability. Finally, extending the approach to multi-stage decision settings and to richer ambiguity set constructions remains an interesting avenue for further investigation.

\section*{Acknowledgments}
This research was supported by Independent Research Fund Denmark, Project No. 2102-00122B.

\bibliographystyle{elsarticle-harv} 

\bibliography{sample}


  



\appendix
\section{Formulations of the Unit Commitment Learning Models}
\label{app:uc_models}
\paragraph{Prescriptive learning}
The prescriptive unit commitment training problem is formulated as
follows:
\begin{subequations}
\label{eq:uc_nominal}
\begin{align}
\min_{\mathbf{w},\mathbf{x},\mathbf{u},\mathbf{p}}
&
\frac{1}{S}
\sum_{s=1}^{S}
\sum_{t=1}^{T}
\sum_{g=1}^{G}
\Big[\!
c_{g}^{U}u_{sgt}
+
c_{g}^{\mathrm{X}}x_{sgt}
+
c_{g}^{P}p_{sgt}+\left|
\mathbf{w}_{gt}^{\top}\mathbf{z}_{st}-x_{sgt}
\right|\!
\Big]
\label{eq:uc_nominal_obj}
\\
\text{s.t.}\quad
&
\mathbf{w}_{gt}^{\top}\mathbf{z}_{st}
\leq
0.5+x_{sgt},
&& \forall s,g,t,
\label{eq:uc_nominal_policy_upper}
\\
&
\mathbf{w}_{gt}^{\top}\mathbf{z}_{st}
\geq
x_{sgt}-0.5,
&& \forall s,g,t,
\label{eq:uc_nominal_policy_lower}
\\
&
-x_{sg(t-1)}+x_{sgt}-x_{sg\tau}
\leq 0,
&& \forall s,g,t,\ \tau\in\{t,\ldots,t+\overline{\nu}_{g}-1\},
\label{eq:uc_nominal_minup}
\\
&
x_{sg(t-1)}-x_{sgt}+x_{sg\tau}-1
\leq 0,
&& \forall s,g,t,\ \tau\in\{t,\ldots,t+\underline{\nu}_{g}-1\},
\label{eq:uc_nominal_mindown}
\\
&
-x_{sg(t-1)}+x_{sgt}-u_{sgt}
\leq 0,
&& \forall s,g,t,
\label{eq:uc_nominal_startup}
\\
&
\underline{p}_{g}x_{sgt}-p_{sgt}
\leq 0,
&& \forall s,g,t,
\label{eq:uc_nominal_pmin}
\\
&
p_{sgt}-\overline{p}_{g}x_{sgt}
\leq 0,
&& \forall s,g,t,
\label{eq:uc_nominal_pmax}
\\
&
p_{sgt}-p_{sg(t-1)}
+
\left(\widetilde{r}_{g}-\overline{r}_{g}\right)
x_{sg(t-1)}
-\widetilde{r}_{g}
\leq 0,
&& \forall s,g,t,
\label{eq:uc_nominal_rampup}
\\
&
p_{sg(t-1)}-p_{sgt}
+
\left(\widetilde{r}_{g}-\underline{r}_{g}\right)
x_{sgt}
-\widetilde{r}_{g}
\leq 0,
&& \forall s,g,t,
\label{eq:uc_nominal_rampdown}
\\
&
\sum_{n=1}^{N}d_{snt}
-
\sum_{g=1}^{G}p_{sgt}
\leq 0,
&& \forall s,t,
\label{eq:uc_nominal_balance}
\\
&
\sum_{n=1}^{N}
h_{\ell n}
\Big(
\sum_{g\in\mathcal{G}_n} p_{sgt}-d_{snt}
\Big)
-\overline{e}_{\ell}
\leq 0,
&& \forall s,\ell,t,
\label{eq:uc_nominal_line_upper}
\\
&
-\sum_{n=1}^{N}
h_{\ell n}
\Big(
\sum_{g\in\mathcal{G}_n} p_{sgt}-d_{snt}
\Big)
-\overline{e}_{\ell}
\leq 0,
&& \forall s,\ell,t,
\label{eq:uc_nominal_line_lower}
\\
&
x_{sgt},u_{sgt}\in\{0,1\},
&& \forall s,g,t.
\label{eq:uc_nominal_domain}
\end{align}
\end{subequations}

Objective~\eqref{eq:uc_nominal_obj} minimizes the average training cost
over all samples. The first three terms
correspond to the physical operating cost of the unit commitment
decisions. Specifically, $c_g^{U}u_{sgt}$ represents the start-up cost
incurred when generator $g$ is switched on, $c_g^{X}x_{sgt}$
represents the fixed online cost of keeping the generator committed, and
$c_g^{P}p_{sgt}$ represents the variable production cost associated
with its power output. The last term,
$\left|\mathbf{w}_{gt}^{\top}\mathbf{z}_{st}-x_{sgt}\right|$, penalizes
the deviation between the continuous output of the learned linear policy
and the corresponding binary commitment decision. This term encourages
the policy output to be consistent with the commitment selected by the
optimization model. Hence, the formulation jointly learns the policy
parameters and the operational decisions, rather than fitting the policy
to precomputed optimal commitment labels.

Constraints~\eqref{eq:uc_nominal_policy_upper} and
\eqref{eq:uc_nominal_policy_lower} link the continuous output of the
linear policy to the binary commitment decision through the threshold
$0.5$. Together with the policy-alignment term in the objective, these
constraints encourage policy values associated with committed units to
lie above the threshold and policy values associated with offline units
to lie below it. The remaining constraints impose the operational requirements of the unit commitment problem~\eqref{eq:uc_deterministic} for every training sample. Specifically, constraints~\eqref{eq:uc_nominal_minup} and \eqref{eq:uc_nominal_mindown} enforce the minimum up- and down-time requirements, respectively; constraint~\eqref{eq:uc_nominal_startup} captures the start-up logic; constraints~\eqref{eq:uc_nominal_pmin} and \eqref{eq:uc_nominal_pmax} impose the minimum and maximum generation limits; constraints~\eqref{eq:uc_nominal_rampup} and \eqref{eq:uc_nominal_rampdown} enforce the ramp-up and ramp-down restrictions; constraint~\eqref{eq:uc_nominal_balance} requires total generation to cover demand; and constraints~\eqref{eq:uc_nominal_line_upper} and \eqref{eq:uc_nominal_line_lower} impose the upper and lower transmission-capacity limits. Finally, constraint~\eqref{eq:uc_nominal_domain} specifies the domains of the binary variables.
\paragraph{Distributionally robust prescriptive learning.}
The distributionally robust model retains the same prescriptive
objective and deterministic intertemporal constraints.
Among the constraints in~\eqref{eq:uc_nominal}, five families contain random
quantities and are therefore replaced by distributionally robust chance
constraints: the upper and lower policy-consistency constraints~\eqref{eq:uc_nominal_policy_upper}--\eqref{eq:uc_nominal_policy_lower}, the power
balance constraint~\eqref{eq:uc_nominal_balance}, and the upper and lower transmission constraints~\eqref{eq:uc_nominal_line_upper}--\eqref{eq:uc_nominal_line_lower}. The
remaining unit commitment constraints are retained in their nominal
sample-wise form.
\paragraph{Policy-consistency constraints.}
Constraint~\eqref{eq:uc_nominal_policy_upper} is written as the individual
distributionally robust chance constraint
\begin{equation}
\inf_{Q\in\mathcal{P}_{\boldsymbol{\omega}}(\rho)}
\mathbb{P}^{Q}
\left[
\mathbf{w}_{gt}^{\top}\mathbf{z}_{t}
-x_{gt}
-0.5
\leq 0
\right]
\geq 1-\epsilon,
\qquad
\forall g,t.
\label{eq:uc_drcc_policy_upper}
\end{equation}
Introducing the auxiliary variables
$q_{sgt}^{+}\in\{0,1\}$, $r_{sgt}^{+}\geq 0$, and
$\widehat{t}_{gt}^{+}\in\mathbb{R}$, the reformulation of
\eqref{eq:uc_drcc_policy_upper} is
\begin{subequations}
\label{eq:uc_reform_policy_upper}
\begin{align}
&\epsilon S\widehat{t}_{gt}^{+}
-
\sum_{s=1}^{S}r_{sgt}^{+}
\geq
\rho S
\left\|
[\mathbf{w}_{gt},-1]
\right\|_{*},
&& \forall g,t,
\label{eq:uc_reform_policy_upper_a}
\\
&-\mathbf{w}_{gt}^{\top}\mathbf{z}_{st}
+x_{sgt}
+0.5
+M_{gt}^{+}q_{sgt}^{+}
\geq
\widehat{t}_{gt}^{+}-r_{sgt}^{+},
&& \forall s,g,t,
\label{eq:uc_reform_policy_upper_b}
\\
&M_{gt}^{+}\left(1-q_{sgt}^{+}\right)
\geq
\widehat{t}_{gt}^{+}-r_{sgt}^{+},
&& \forall s,g,t,
\label{eq:uc_reform_policy_upper_c}
\\
&q_{sgt}^{+}\in\{0,1\},
\quad
r_{sgt}^{+}\geq 0,
\quad
\widehat{t}_{gt}^{+}\in\mathbb{R},
&& \forall s,g,t.
\label{eq:uc_reform_policy_upper_d}
\end{align}
\end{subequations}
Constraint~\eqref{eq:uc_nominal_policy_lower} is equivalently expressed as
\begin{equation}
\inf_{Q\in\mathcal{P}_{\boldsymbol{\omega}}(\rho)}
\mathbb{P}^{Q}
\left[
-\mathbf{w}_{gt}^{\top}\mathbf{z}_{t}
+x_{gt}
-0.5
\leq 0
\right]
\geq 1-\epsilon,
\qquad
\forall g,t.
\label{eq:uc_drcc_policy_lower}
\end{equation}
Introducing
$q_{sgt}^{-}\in\{0,1\}$, $r_{sgt}^{-}\geq 0$, and
$\widehat{t}_{gt}^{-}\in\mathbb{R}$, the reformulation of~\eqref{eq:uc_drcc_policy_lower} gives
\begin{subequations}
\label{eq:uc_reform_policy_lower}
\begin{align}
&\epsilon S\widehat{t}_{gt}^{-}
-
\sum_{s=1}^{S}r_{sgt}^{-}
\geq
\rho S
\left\|
[-\mathbf{w}_{gt},1]
\right\|_{*},
&& \forall g,t,
\label{eq:uc_reform_policy_lower_a}
\\
&\mathbf{w}_{gt}^{\top}\mathbf{z}_{st}
-x_{sgt}
+0.5
+M_{gt}^{-}q_{sgt}^{-}
\geq
\widehat{t}_{gt}^{-}-r_{sgt}^{-},
&& \forall s,g,t,
\label{eq:uc_reform_policy_lower_b}
\\
&M_{gt}^{-}\left(1-q_{sgt}^{-}\right)
\geq
\widehat{t}_{gt}^{-}-r_{sgt}^{-},
&& \forall s,g,t,
\label{eq:uc_reform_policy_lower_c}
\\
&q_{sgt}^{-}\in\{0,1\},
\quad
r_{sgt}^{-}\geq 0,
\quad
\widehat{t}_{gt}^{-}\in\mathbb{R},
&& \forall s,g,t.
\label{eq:uc_reform_policy_lower_d}
\end{align}
\end{subequations}
\paragraph{Power balance constraint.}
The power balance constraint~\eqref{eq:uc_nominal_balance} becomes
\begin{equation}
\inf_{Q\in\mathcal{P}_{\boldsymbol{\omega}}(\rho)}
\mathbb{P}^{Q}
\left[
\sum_{n=1}^{N}d_{nt}
-
\sum_{g=1}^{G}p_{gt}
\leq 0
\right]
\geq 1-\epsilon,
\qquad
\forall t.
\label{eq:uc_drcc_balance}
\end{equation}
Introducing
$q_{st}^{\mathrm{bal}}\in\{0,1\}$,
$r_{st}^{\mathrm{bal}}\geq 0$, and
$\widehat{t}_{t}^{\mathrm{bal}}\in\mathbb{R}$, the reformulation of~\eqref{eq:uc_drcc_balance} is
\begin{subequations}
\label{eq:uc_reform_balance}
\begin{align}
&\epsilon S\widehat{t}_{t}^{\mathrm{bal}}
-
\sum_{s=1}^{S}r_{st}^{\mathrm{bal}}
\geq
\rho S(N+G),
&& \forall t,
\label{eq:uc_reform_balance_a}
\\
&\sum_{g=1}^{G}p_{sgt}
-
\sum_{n=1}^{N}d_{snt}
+
M_{t}^{\mathrm{bal}}q_{st}^{\mathrm{bal}}
\geq
\widehat{t}_{t}^{\mathrm{bal}}
-
r_{st}^{\mathrm{bal}},
&& \forall s,t,
\label{eq:uc_reform_balance_b}
\\
&M_{t}^{\mathrm{bal}}
\left(
1-q_{st}^{\mathrm{bal}}
\right)
\geq
\widehat{t}_{t}^{\mathrm{bal}}
-
r_{st}^{\mathrm{bal}},
&& \forall s,t,
\label{eq:uc_reform_balance_c}
\\
&q_{st}^{\mathrm{bal}}\in\{0,1\},
\quad
r_{st}^{\mathrm{bal}}\geq 0,
\quad
\widehat{t}_{t}^{\mathrm{bal}}\in\mathbb{R},
&& \forall s,t.
\label{eq:uc_reform_balance_d}
\end{align}
\end{subequations}
\paragraph{Upper transmission constraint.}
The upper transmission constraint~\eqref{eq:uc_nominal_line_upper} is replaced by
\begin{equation}
\inf_{Q\in\mathcal{P}_{\boldsymbol{\omega}}(\rho)}
\mathbb{P}^{Q}
\left[
\sum_{n=1}^{N}
h_{\ell n}
\Big(
\sum_{g\in\mathcal{G}_n} p_{gt}-d_{nt}
\Big)
-\overline{e}_{\ell}
\leq 0
\right]
\geq 1-\epsilon,
\qquad
\forall \ell,t.
\label{eq:uc_drcc_line_upper}
\end{equation}
Let $n(g)$ denote the bus at which generating unit $g$ is located, so that
$\sum_{n=1}^{N}h_{\ell n}\sum_{g\in\mathcal{G}_n}p_{gt}=\sum_{g=1}^{G}h_{\ell,n(g)}p_{gt}$.
Introducing
$q_{s\ell t}^{+}\in\{0,1\}$,
$r_{s\ell t}^{+}\geq 0$, and
$\widehat{t}_{\ell t}^{+}\in\mathbb{R}$, the reformulation of~\eqref{eq:uc_drcc_line_upper} gives
\begin{subequations}
\label{eq:uc_reform_line_upper}
\begin{align}
&\epsilon S\widehat{t}_{\ell t}^{+}
-
\sum_{s=1}^{S}r_{s\ell t}^{+}
\geq
\rho S
\Big(
\sum_{g=1}^{G}
\left|
h_{\ell,n(g)}
\right|
+
\sum_{n=1}^{N}
\left|
h_{\ell n}
\right|
\Big),
&& \forall \ell,t,
\label{eq:uc_reform_line_upper_a}
\\
&\overline{e}_{\ell}
+
\sum_{n=1}^{N}h_{\ell n}d_{snt}
-
\sum_{g=1}^{G}h_{\ell,n(g)}p_{sgt}
+
M_{\ell t}^{+}q_{s\ell t}^{+}
\geq
\widehat{t}_{\ell t}^{+}
-
r_{s\ell t}^{+},
&& \forall s,\ell,t,
\label{eq:uc_reform_line_upper_b}
\\
&M_{\ell t}^{+}
\left(
1-q_{s\ell t}^{+}
\right)
\geq
\widehat{t}_{\ell t}^{+}
-
r_{s\ell t}^{+},
&& \forall s,\ell,t,
\label{eq:uc_reform_line_upper_c}
\\
&q_{s\ell t}^{+}\in\{0,1\},
\quad
r_{s\ell t}^{+}\geq 0,
\quad
\widehat{t}_{\ell t}^{+}\in\mathbb{R},
&& \forall s,\ell,t.
\label{eq:uc_reform_line_upper_d}
\end{align}
\end{subequations}
\paragraph{Lower transmission constraint.}
The lower transmission constraint~\eqref{eq:uc_nominal_line_lower} becomes
\begin{equation}
\inf_{Q\in\mathcal{P}_{\boldsymbol{\omega}}(\rho)}
\mathbb{P}^{Q}
\left[
-\sum_{n=1}^{N}
h_{\ell n}
\Big(
\sum_{g\in\mathcal{G}_n} p_{gt}-d_{nt}
\Big)
-\overline{e}_{\ell}
\leq 0
\right]
\geq 1-\epsilon,
\qquad
\forall \ell,t.
\label{eq:uc_drcc_line_lower}
\end{equation}
Introducing
$q_{s\ell t}^{-}\in\{0,1\}$,
$r_{s\ell t}^{-}\geq 0$, and
$\widehat{t}_{\ell t}^{-}\in\mathbb{R}$, the reformulation of~\eqref{eq:uc_drcc_line_lower} is
\begin{subequations}
\label{eq:uc_reform_line_lower}
\begin{align}
&\epsilon S\widehat{t}_{\ell t}^{-}
-
\sum_{s=1}^{S}r_{s\ell t}^{-}
\geq
\rho S
\Big(
\sum_{g=1}^{G}
\left|
h_{\ell,n(g)}
\right|
+
\sum_{n=1}^{N}
\left|
h_{\ell n}
\right|
\Big),
&& \forall \ell,t,
\label{eq:uc_reform_line_lower_a}
\\
&\overline{e}_{\ell}
-
\sum_{n=1}^{N}h_{\ell n}d_{snt}
+
\sum_{g=1}^{G}h_{\ell,n(g)}p_{sgt}
+
M_{\ell t}^{-}q_{s\ell t}^{-}
\geq
\widehat{t}_{\ell t}^{-}
-
r_{s\ell t}^{-},
&& \forall s,\ell,t,
\label{eq:uc_reform_line_lower_b}
\\
&M_{\ell t}^{-}
\left(
1-q_{s\ell t}^{-}
\right)
\geq
\widehat{t}_{\ell t}^{-}
-
r_{s\ell t}^{-},
&& \forall s,\ell,t,
\label{eq:uc_reform_line_lower_c}
\\
&q_{s\ell t}^{-}\in\{0,1\},
\quad
r_{s\ell t}^{-}\geq 0,
\quad
\widehat{t}_{\ell t}^{-}\in\mathbb{R},
&& \forall s,\ell,t.
\label{eq:uc_reform_line_lower_d}
\end{align}
\end{subequations}
Combining objective~\eqref{eq:uc_nominal_obj}, the deterministic operational
constraints~\eqref{eq:uc_nominal_minup}--\eqref{eq:uc_nominal_rampdown} and~\eqref{eq:uc_nominal_domain}, and the
distributionally robust reformulations~\eqref{eq:uc_reform_policy_upper}, \eqref{eq:uc_reform_policy_lower}, \eqref{eq:uc_reform_balance}, \eqref{eq:uc_reform_line_upper}, and~\eqref{eq:uc_reform_line_lower}
yields the distributionally robust prescriptive unit commitment model.

\section{Formulations of the Staff Allocation Learning Models}
\label{app:sa_models}
\paragraph{Prescriptive learning}
We compare classical predictive learning, prescriptive learning,
and distributionally robust prescriptive learning.
The prescriptive staff allocation training problem is formulated as
follows:
\begin{subequations}
\label{eq:sa_nominal}
\begin{align}
\min_{\mathbf{w},\mathbf{x}}
\quad
&
\frac{1}{S}
\sum_{s=1}^{S}
\left[
\sum_{e=1}^{E}\sum_{j=1}^{J}
c_{e}\,x_{sej}
+
\sum_{e=1}^{E}\sum_{j=1}^{J}
\left|\mathbf{w}_{ej}^{\top}\mathbf{z}_{s}-x_{sej}\right|
\right]
\label{eq:sa_nominal_obj}
\\
\text{s.t.}\quad
&
\mathbf{w}_{ej}^{\top}\mathbf{z}_{s}
\leq
0.5+x_{sej},
&& \forall s,e,j,
\label{eq:sa_nominal_policy_upper}
\\
&
\mathbf{w}_{ej}^{\top}\mathbf{z}_{s}
\geq
x_{sej}-0.5,
&& \forall s,e,j,
\label{eq:sa_nominal_policy_lower}
\\
&
\sum_{e=1}^{E} x_{sej}
\geq d_{sj},
&& \forall s,j,
\label{eq:sa_nominal_demand}
\\
&
\underline{s}_{e}
\leq
\sum_{j=1}^{J} x_{sej}
\leq \overline{s}_{e},
&& \forall s,e,
\label{eq:sa_nominal_shiftlimits}
\\
&
x_{sej}\in\{0,1\},
&& \forall s,e,j.
\label{eq:sa_nominal_domain}
\end{align}
\end{subequations}
Objective~\eqref{eq:sa_nominal_obj} minimizes the average total staffing
cost across all training samples, augmented by a penalty
$\left|\mathbf{w}_{ej}^{\top}\mathbf{z}_{s}-x_{sej}\right|$
that encourages the continuous policy output to align with the binary
assignment decision.
Constraints~\eqref{eq:sa_nominal_policy_upper} and
\eqref{eq:sa_nominal_policy_lower} enforce that the linear policy score
$\mathbf{w}_{ej}^{\top}\mathbf{z}_{s}$ lies within a symmetric band of
$\pm 0.5$ around the binary decision $x_{sej}$, effectively linking the
learned policy to the assignment variable through a soft rounding
mechanism. Constraint~\eqref{eq:sa_nominal_demand} ensures that each
work shift $j$ is covered by at least $d_{sj}$ employees for every training
sample $s$. Constraint~\eqref{eq:sa_nominal_shiftlimits} imposes the
individual workload limits for every training sample.

\paragraph{Distributionally robust prescriptive learning.}
The distributionally robust model retains the same prescriptive objective
and the nominal workload-limit constraints~\eqref{eq:sa_nominal_shiftlimits}, since the parameters
$\underline{s}_{e}$ and $\overline{s}_{e}$ are deterministic. Two
families of constraints in~\eqref{eq:sa_nominal} contain random
quantities---the policy-consistency constraints~\eqref{eq:sa_nominal_policy_upper}--\eqref{eq:sa_nominal_policy_lower} and the demand coverage
constraint~\eqref{eq:sa_nominal_demand}---and are therefore replaced by their Wasserstein
distributionally robust chance constraint counterparts.
\paragraph{Policy-consistency constraints.}
Constraint~\eqref{eq:sa_nominal_policy_upper} is replaced by
the individual distributionally robust chance constraint
\begin{equation}
\inf_{Q\in\mathcal{P}_{\boldsymbol{\omega}}(\rho)}
\mathbb{P}^{Q}
\left[
\mathbf{w}_{ej}^{\top}\mathbf{z}
-x_{ej}
-0.5
\leq 0
\right]
\geq 1-\epsilon,
\qquad
\forall e,j.
\label{eq:sa_drcc_policy_upper}
\end{equation}
Introducing the auxiliary variables
$q_{sej}^{+}\in\{0,1\}$, $r_{sej}^{+}\geq 0$, and
$\widehat{t}_{ej}^{+}\in\mathbb{R}$, the reformulation
of~\eqref{eq:sa_drcc_policy_upper} is
\begin{subequations}
\label{eq:sa_reform_policy_upper}
\begin{align}
&\epsilon S\,\widehat{t}_{ej}^{+}
-
\sum_{s=1}^{S}r_{sej}^{+}
\geq
\rho S
\left\|
[\mathbf{w}_{ej},-1]
\right\|_{*},
&& \forall e,j,
\label{eq:sa_reform_pu_a}
\\
&-\mathbf{w}_{ej}^{\top}\mathbf{z}_{s}
+x_{sej}
+0.5
+M_{ej}^{+}q_{sej}^{+}
\geq
\widehat{t}_{ej}^{+}-r_{sej}^{+},
&& \forall s,e,j,
\label{eq:sa_reform_pu_b}
\\
&M_{ej}^{+}\left(1-q_{sej}^{+}\right)
\geq
\widehat{t}_{ej}^{+}-r_{sej}^{+},
&& \forall s,e,j,
\label{eq:sa_reform_pu_c}
\\
&q_{sej}^{+}\in\{0,1\},
\quad
r_{sej}^{+}\geq 0,
\quad
\widehat{t}_{ej}^{+}\in\mathbb{R},
&& \forall s,e,j.
\label{eq:sa_reform_pu_d}
\end{align}
\end{subequations}

Constraint~\eqref{eq:sa_nominal_policy_lower} is equivalently
written as the distributionally robust chance constraint
\begin{equation}
\inf_{Q\in\mathcal{P}_{\boldsymbol{\omega}}(\rho)}
\mathbb{P}^{Q}
\left[
-\mathbf{w}_{ej}^{\top}\mathbf{z}
+x_{ej}
-0.5
\leq 0
\right]
\geq 1-\epsilon,
\qquad
\forall e,j.
\label{eq:sa_drcc_policy_lower}
\end{equation}
Introducing $q_{sej}^{-}\in\{0,1\}$, $r_{sej}^{-}\geq 0$, and
$\widehat{t}_{ej}^{-}\in\mathbb{R}$, the reformulation of~\eqref{eq:sa_drcc_policy_lower} is
\begin{subequations}
\label{eq:sa_reform_policy_lower}
\begin{align}
&\epsilon S\,\widehat{t}_{ej}^{-}
-
\sum_{s=1}^{S}r_{sej}^{-}
\geq
\rho S
\left\|
[-\mathbf{w}_{ej},1]
\right\|_{*},
&& \forall e,j,
\label{eq:sa_reform_pl_a}
\\
&\mathbf{w}_{ej}^{\top}\mathbf{z}_{s}
-x_{sej}
+0.5
+M_{ej}^{-}q_{sej}^{-}
\geq
\widehat{t}_{ej}^{-}-r_{sej}^{-},
&& \forall s,e,j,
\label{eq:sa_reform_pl_b}
\\
&M_{ej}^{-}\left(1-q_{sej}^{-}\right)
\geq
\widehat{t}_{ej}^{-}-r_{sej}^{-},
&& \forall s,e,j,
\label{eq:sa_reform_pl_c}
\\
&q_{sej}^{-}\in\{0,1\},
\quad
r_{sej}^{-}\geq 0,
\quad
\widehat{t}_{ej}^{-}\in\mathbb{R},
&& \forall s,e,j.
\label{eq:sa_reform_pl_d}
\end{align}
\end{subequations}

\paragraph{Demand coverage constraint.}
Constraint~\eqref{eq:sa_nominal_demand} is
replaced by the distributionally robust chance constraint
\begin{equation}
\inf_{Q\in\mathcal{P}_{\boldsymbol{\omega}}(\rho)}
\mathbb{P}^{Q}
\left[
-\sum_{e=1}^{E}x_{ej}+d_{j}
\leq 0
\right]
\geq 1-\epsilon,
\qquad
\forall j.
\label{eq:sa_drcc_demand}
\end{equation}
Introducing $q_{sj}^{\mathrm{dem}}\in\{0,1\}$, $r_{sj}^{\mathrm{dem}}\geq 0$, and
$\widehat{t}_{j}^{\mathrm{dem}}\in\mathbb{R}$, the reformulation
of~\eqref{eq:sa_drcc_demand} is
\begin{subequations}
\label{eq:sa_reform_demand}
\begin{align}
&\epsilon S\,\widehat{t}_{j}^{\mathrm{dem}}
-
\sum_{s=1}^{S}r_{sj}^{\mathrm{dem}}
\geq
\rho S\,(1+E),
&& \forall j,
\label{eq:sa_reform_dem_a}
\\
&\sum_{e=1}^{E}x_{sej}
-d_{sj}
+M_{j}^{\mathrm{dem}}q_{sj}^{\mathrm{dem}}
\geq
\widehat{t}_{j}^{\mathrm{dem}}-r_{sj}^{\mathrm{dem}},
&& \forall s,j,
\label{eq:sa_reform_dem_b}
\\
&M_{j}^{\mathrm{dem}}\left(1-q_{sj}^{\mathrm{dem}}\right)
\geq
\widehat{t}_{j}^{\mathrm{dem}}-r_{sj}^{\mathrm{dem}},
&& \forall s,j,
\label{eq:sa_reform_dem_c}
\\
&q_{sj}^{\mathrm{dem}}\in\{0,1\},
\quad
r_{sj}^{\mathrm{dem}}\geq 0,
\quad
\widehat{t}_{j}^{\mathrm{dem}}\in\mathbb{R},
&& \forall s,j.
\label{eq:sa_reform_dem_d}
\end{align}
\end{subequations}
The resulting distributionally robust prescriptive learning model
minimizes objective~\eqref{eq:sa_nominal_obj} subject to the workload
limits~\eqref{eq:sa_nominal_shiftlimits}, the domain
constraint~\eqref{eq:sa_nominal_domain}, and the
reformulations~\eqref{eq:sa_reform_policy_upper}, \eqref{eq:sa_reform_policy_lower}, and~\eqref{eq:sa_reform_demand}.

\end{document}